\documentclass[5p,times]{elsarticle}

\usepackage{hyperref}
\usepackage{subcaption}
\usepackage{graphicx}
\usepackage{xurl}
\usepackage{nomencl}
\makenomenclature

\journal{Energy and Buildings}

\begin{document}

\begin{frontmatter}

\title{Creating synthetic energy meter data using conditional diffusion and building metadata}

\author[inst1]{Chun Fu}
\author[inst2]{Hussain Kazmi}
\author[inst1]{Matias Quintana}
\author[inst1]{Clayton Miller\corref{cor1}}

\ead{clayton@nus.edu.sg}
\cortext[cor1]{Corresponding author}

\affiliation[inst1]{organization={Building and Urban Data Science (BUDS) Lab, Department of the Built Environment, College of Design and Engineering, National University of Singapore (NUS)},
            country={Singapore}}
\affiliation[inst2]{organization={Department of Electrical Engineering (ELECTA-ESAT), KU Leuven}, country={Belgium}}

\begin{abstract}
Advances in machine learning and increased computational power have driven progress in energy-related research. However, limited access to private energy data from buildings hinders traditional regression models relying on historical data. While generative models offer a solution, previous studies have primarily focused on short-term generation periods (e.g., daily profiles) and a limited number of meters.
Thus, the study proposes a conditional diffusion model for generating high-quality synthetic energy data using relevant metadata. Using a dataset comprising 1,828 power meters from various buildings and countries, this model is compared with traditional methods like Conditional Generative Adversarial Networks (CGAN) and Conditional Variational Auto-Encoders (CVAE). It explicitly handles long-term annual consumption profiles, harnessing metadata such as location, weather, building, and meter type to produce coherent synthetic data that closely resembles real-world energy consumption patterns.
The results demonstrate the proposed diffusion model's superior performance, with a 36\% reduction in Fr'echet Inception Distance (FID) score and a 13\% decrease in Kullback-Leibler divergence (KL divergence) compared to the following best method. The proposed method successfully generates high-quality energy data through metadata, and its code will be open-sourced, establishing a foundation for a broader array of energy data generation models in the future.
\end{abstract}


\end{frontmatter}

\nomenclature{ML}{Machine Learning}
\nomenclature{AI}{Artificial Intelligence}
\nomenclature{BDG2}{Building Data Genome 2.0}

\nomenclature{FDD}{Fault Detection and Diagnosis}
\nomenclature{GAN}{Generative Adversarial Network}
\nomenclature{CGAN}{Conditional Generative Adversarial Network}
\nomenclature{CVAE}{Conditional Variational Autoencoder}
\nomenclature{VAE}{Variational Autoencoder}
\nomenclature{BPS}{Building Performance Simulation}
\nomenclature{LSTM}{Long Short-Term Memory}
\nomenclature{ReLU}{Rectified Linear Unit}
\nomenclature{RMSE}{Root Mean Squared Error}
\nomenclature{MAE}{Mean Absolute Error}
\nomenclature{MSE}{Mean Squared Error}

\nomenclature{R\textsuperscript{2}}{Coefficient of Determination}
\nomenclature{FID}{Fréchet Inception Distance}
\nomenclature{KL}{Kullback-Leibler}

\printnomenclature


\section{Introduction}

The ongoing processes of urbanization and industrialization continue to drive global energy demand, underscoring the importance of research into energy consumption and forecasting \cite{creutzig2015global}. The increasing digitization and interconnectivity of our societies have elevated the significance of energy, with its efficient utilization serving as a fundamental element of technological progress and economic development \cite{stephenson2010energy}. Notably, the adoption of smart meters has witnessed remarkable growth, with approximately 729.1 million smart meters globally in 2019, representing a staggering decade-long growth rate of 3013\% since 2010 \cite{sovacool2021global}. Furthermore, predictions from another study suggest that the global market for smart electricity meters, initially estimated at US \$10.5 billion in 2020, is projected to experience an annual growth rate of 6.7\%, reaching a valuation of USD \$15.2 billion by 2026 \cite{arora2022review}. 

Within the European Union (EU), member countries are mandated to deploy a minimum of 80\% smart meters by 2020 \cite{union2009directive}, and approximately 75\% of U.S. households are estimated to have adopted smart meters \cite{cooper2021electric}. However, despite the widespread adoption of smart meters, data scarcity persists due to legitimate privacy and security concerns \cite{mcdaniel2009security, balta2013social}. There are also concerns about the potential identification of consumer profiles through high-frequency readings \cite{langer2013privacy}. Consequently, despite having an increasing number of installed meters and the data collected from them, energy companies and grid operators remain reluctant or are not allowed to share energy data. This situation creates a paradoxical scenario where data remains insufficient despite the exponential growth in the number of meters \cite{kazmi2021towards}. As a result, data insufficiency, particularly in light of frequent anomalies and missing energy data, is one of the primary challenges that must be addressed.


\subsection{Conventional regression-based energy model and building performance simulation}
The realm of energy forecasting has entered a transformative phase due to continuous advancements in computational capabilities, innovative data processing techniques, and the emergence of novel algorithmic approaches \cite{kazmi2023ten}. Machine learning (ML) technologies have demonstrated considerable potential in predicting energy consumption in buildings \cite{somu2021deep, pham2020predicting, olu2022building}, forecasting renewable energy output \cite{dolara2017weather, gensler2016deep, tian2022developing}, and estimating grid electricity demand \cite{hafeez2020novel, avalos2020comparative, hafeez2020electric}.

Conventional regression-based energy models are frequently employed for energy forecasting tasks, aiming to establish connections between energy consumption and various independent variables \cite{fumo2015regression}. These models span a spectrum of complexity, encompassing simple linear regression to more sophisticated techniques like support vector machines and neural networks \cite{ahmad2020review}. 
Based on a survey conducted among the winning teams of the Great Energy Predictor III competition, it was found that over 70\% of respondents favored gradient boosting regression models, such as XGBoost, LightGBM, and CatBoost, as their preferred algorithm for forecasting energy consumption \cite{Miller2022-jj}.
While these models provide reasonably accurate predictions, a significant drawback of regression-based approaches is their substantial data requirements, often mandating months of historical readings to capture seasonal and behavioral patterns adequately \cite{kazmi2019multi}. To ensure comprehensive data representation, many studies rely on extensive datasets spanning months or even years \cite{amasyali2018review}. This limitation hampers their applicability in scenarios with limited meter data availability.

Another conventional approach is Building Performance Simulation (BPS) tools, which create virtual representations of buildings to estimate energy usage profiles under various conditions \cite{hensen2012building, hong2020generation}. However, the development of accurate BPS models demands meticulous audits to capture precise architectural and operational details, as well as a laborious calibration process to minimize discrepancies \cite{reddy2006literature, fabrizio2015}. This intensive effort and expertise required can often render the widespread adoption of BPS impractical. Thus, while traditional regression and simulation methods have contributed to energy analysis, their heavy reliance on data and parameters presents limitations.

\subsection{Emergence of generative models and their application in the energy field}

In contrast to traditional approaches, the advent of generative models has sparked revolutionary changes in predictive modeling \cite{goodfellow2014generative}. Among the various ML paradigms, generative models have shown a capacity to generate new data that mimic the distribution of training data, showing promise in a variety of fields, including energy data analysis. Models such as Generative Adversarial Networks (GANs) have been increasingly adopted to learn the distribution of the input data and generate synthetic data resembling the original data.

Recent advances in generative models, such as GANs, Variational Auto-Encoders (VAEs) \cite{kingma2013auto}, and diffusion models\cite{ho2020denoising}, offer a potential solution to the issue of data scarcity.
These models have demonstrated the capacity to create new data that resembles the original dataset, becoming increasingly useful in various fields. They have shown particular promise in applications such as power demand prediction \cite{yeEnergyBuildingsEvaluating2022}, building load generation \cite{chen2018model, el2020data, wang2020generating}, Fault Detection and Diagnostics (FDD) \cite{yan2020generative}, meter classification \cite{fu2023enhancing}, and indoor thermal comfort \cite{quintana2020balancing}. 
However, they are not without their drawbacks. 
Most existing generative research focuses on short-term energy data, primarily daily profiles, leaving a significant gap in long-term data applications. Privacy concerns and data scarcity further exacerbate these challenges, often limiting the availability of energy data from individual buildings and community power grids. Limited datasets, such as from a single power meter or site, can result in models that lack generalizability. Moreover, many of these generative models in past research do not include conditions related to building characteristics from metadata, like building and meter types. This lack of specificity hinders the models' ability to generate energy data under particular conditions, compromising their flexibility and variability.


\subsection{Integration of meta-information into generative models}

Recent studies have explored the incorporation of meaningful metadata or side information to guide generative models in synthesizing more realistic and meaningful data samples \cite{mirza2014conditional}. Unlike unconditional models, which generate data from latent variables, conditional models allow additional contextual variables to direct the generation process. In computer vision, conditioning on class labels enables controllable image generation adhering to specified categories. For sequential data like text or audio, linguistic features can be provided as conditions to preserve semantic coherence \cite{ramponi2018t, fu2019time,guo2018long}.

In the building energy domain, metadata attributes offer valuable contextual cues for energy data synthesis. 
A conditional GAN model was incorporated with mean monthly outdoor temperature into the generative process and proved that the additional condition could improve the performance \cite{baasch2021conditional}. 
A conditional TimeGAN, employing recurrent architecture and capable of incorporating multiple input time series such as temperature and solar radiation, was introduced to generate energy data aimed at improving subsequent reinforcement learning task \cite{fochesato2022use}.
Some other research efforts have shown the successful synthesis of energy data that effectively integrates meta-information \cite{salatiello2023synthesizing, nuastuasescu2022conditional}.
Nevertheless, the integration of such meta-information into generative models poses challenges, necessitating a delicate balance between model complexity and the accuracy of the generated data \cite{park2018data}. Excessively complex models may lead to overfitting, while overly simplified models might yield less useful synthetic data. 
Consequently, issues related to the quality and training of GANs for synthetic energy data have persisted, restricting their application to daily profiles or monthly data. 
Despite these challenges, recent advancements in diffusion models have addressed common issues encountered in GANs, providing more stable training processes and yielding high-quality generated data, thereby unlocking promising applications \cite{croitoru2023diffusion}.

Moreover, it is noteworthy that prior studies have not integrated building metadata as conditions for generating energy data. Information about building types, including schools, offices, or residences, offers valuable insights into occupancy and usage patterns, which exert a substantial influence on energy consumption \cite{fu2022using}. Additionally, knowledge of installed meter types, whether electricity, gas, or water-based, indicates the form of energy usage being measured. Geographic location also plays a vital role in determining climate impacts and solar availability, which affect building energy profiles. Generative models can produce more accurate, variable energy consumption patterns aligned with real-world characteristics by incorporating such informative metadata.

By integrating such meta-information, we pave the way for more targeted energy efficiency initiatives and personalized energy management strategies. Building upon these efforts, our study introduces a conditional diffusion model driven by metadata. This meta-driven diffusion model is designed to generate synthetic long-term annual load data, tackling the limitations inherent in traditional energy consumption forecasting and enhancing the data availability for energy management tasks. The effective incorporation of this wealth of meta-information into generative models for synthetic data generation remains a complex and relatively unexplored area, calling for further research. The study presented in this paper aims to take a step in this direction.

\subsection{Research Objectives and Novelty}

This study embarks on a path to advance the generation of synthetic energy data by introducing a novel approach based on conditional generative models. We aim to address the limitations in current practices and contribute to the state of the art in this critical research area. Our research objectives and the novelty of this work can be summarized as follows:

\begin{enumerate}
    \item \textbf{Meta-driven conditional generative modeling for energy data:} 
    
    The primary objective of this study is to implement state-of-the-art conditional generative models for energy data generation. These models would be able to generate synthetic energy data based on specified conditions, reflecting the influence of metadata like meter types and building types on energy consumption.
    
    \item \textbf{Long-term energy data generation via computer vision algorithms:} 
    
    Unlike most studies that focus on short-term energy data, this study aims to generate synthetic energy data spanning a full year. This addresses a significant gap in the literature, providing a long-term, high-resolution synthetic dataset for more accurate and comprehensive energy forecasting. This is realized by integrating deep learning generative models derived from the computer vision domain with reshaped energy data, enabling the model to grasp the underlying patterns of energy consumption efficiently.
    
    \item \textbf{Validation of model performance and applicability on an open energy consumption dataset across buildings and countries:}
    
    In this research, we will evaluate the effectiveness of our developed generative models using the extensive BDG2 dataset, which comprises power meters from around the world. This will enable us to evaluate the models' ability to adapt to the wide-ranging characteristics of energy data. Our evaluation metrics will encompass various metrics, including KL divergence and FID, to assess the distribution similarity and diversity of the generated results, thus providing a comprehensive analysis of the generative models. 
    
\end{enumerate}

\begin{figure*}[!htb]
\begin{center}
\includegraphics[width=0.8\textwidth, trim= 0cm 0cm 0cm 0cm,clip]{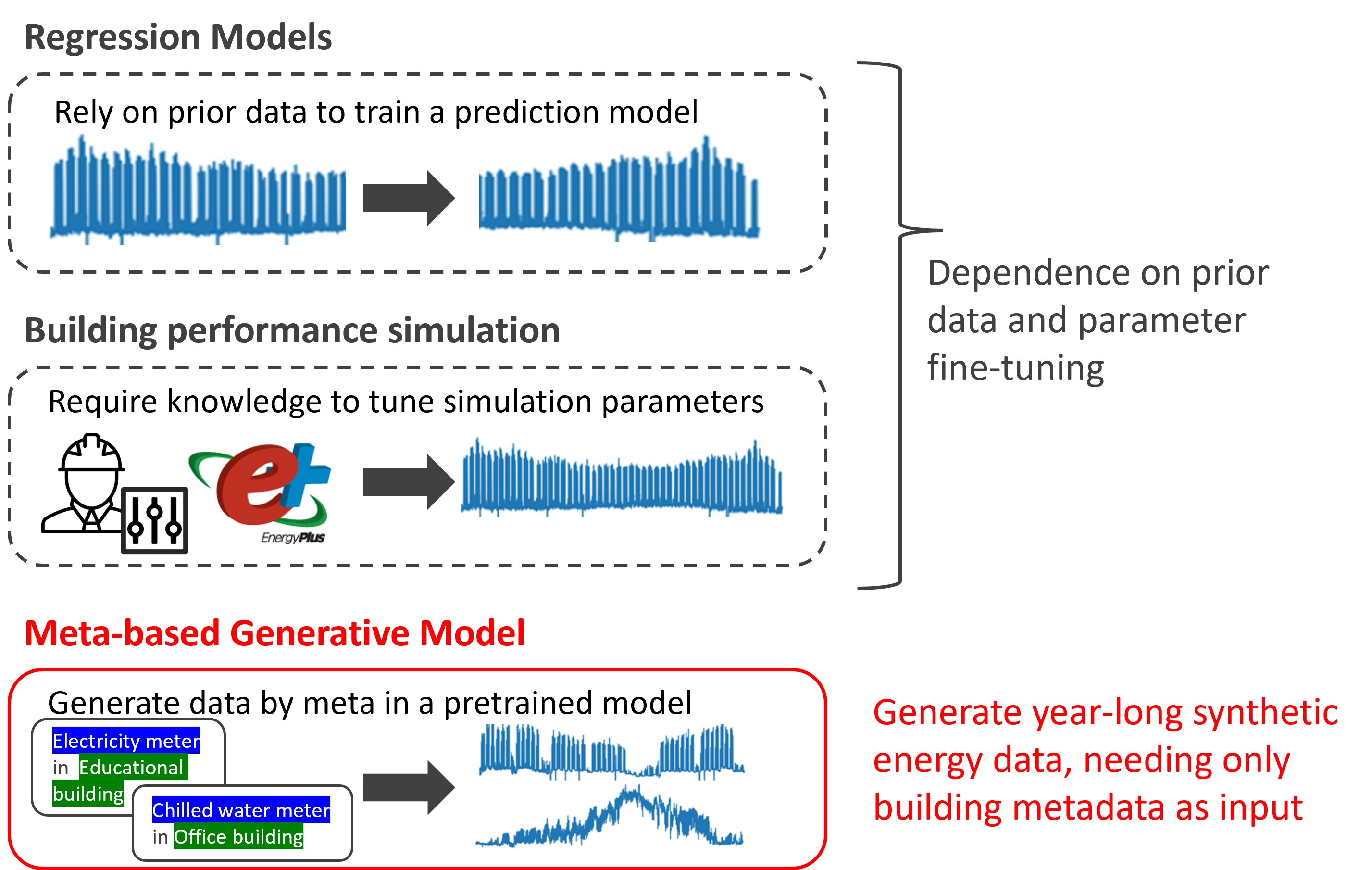}
\caption{Comparison between traditional methods and our proposed meta-driven generative model.}
\label{fig:concept}
\end{center}
\end{figure*}

The study introduces a comprehensive framework that seamlessly integrates meta-information into conditional generative models. This reduces dependency on abundant historical data and eliminates the need for laborious parameter tuning, which is a challenge commonly faced with traditional methods such as regression models and building performance simulation (BPS). These improvements over traditional techniques are visually summarized in \autoref{fig:concept}. Furthermore, by emphasizing the generation of long-term, high-resolution data, the study opens new avenues for the broader application of generative models in the energy field for the future.

\section{Methodology}

\subsection{Dataset: Building Data Genome 2.0 (BDG2.0)}
The present research employs hourly time-series data from electrical meters, sourced from the Building Data Genome 2.0 (BDG2) project \cite{miller2020building}, for modeling purposes. The BDG2 dataset is publicly accessible and comprises hourly readings from 3,053 meters, collected over a two-year period. Given its diverse range of meter data from various geographic locations, the BDG2 dataset serves as an ideal benchmark for evaluating the effectiveness and adaptability of different machine learning models. 
Table \ref{tab:bdg2_feat} provides an overview of the BDG2 dataset and its contained metadata variables.

\begin{table}[!htb]
\centering
\resizebox{0.5\textwidth}{!}{%
\begin{tabular}{ll}
\hline
\textbf{Feature} & \textbf{Details} \\ \hline
Dataset Source & Building Data Genome 2.0 Project \\
Time Range & 2016 and 2017 (2 years) \\
Frequency & Hourly \\
Total Meters & 3053 \\
Total Buildings & 1636 \\
Meter Types & \begin{tabular}[c]{@{}l@{}}Electricity, Chilled water, Steam,  Hot water, \\ Gas, Water, Irrigation, and Solar\end{tabular} \\
Building Types & \begin{tabular}[c]{@{}l@{}}Education, Office, Entertainment/public assembly, \\ Lodging/residential, and Public services\end{tabular} \\ \hline
\end{tabular}
}
\caption{Main features of the dataset sourced from the Building Data Genome 2.0 Project.}
\label{tab:bdg2_feat}
\end{table}

\begin{figure*}[!htb]
\begin{center}
\includegraphics[width=0.95\textwidth, trim= 0cm 0cm 0cm 0cm,clip]{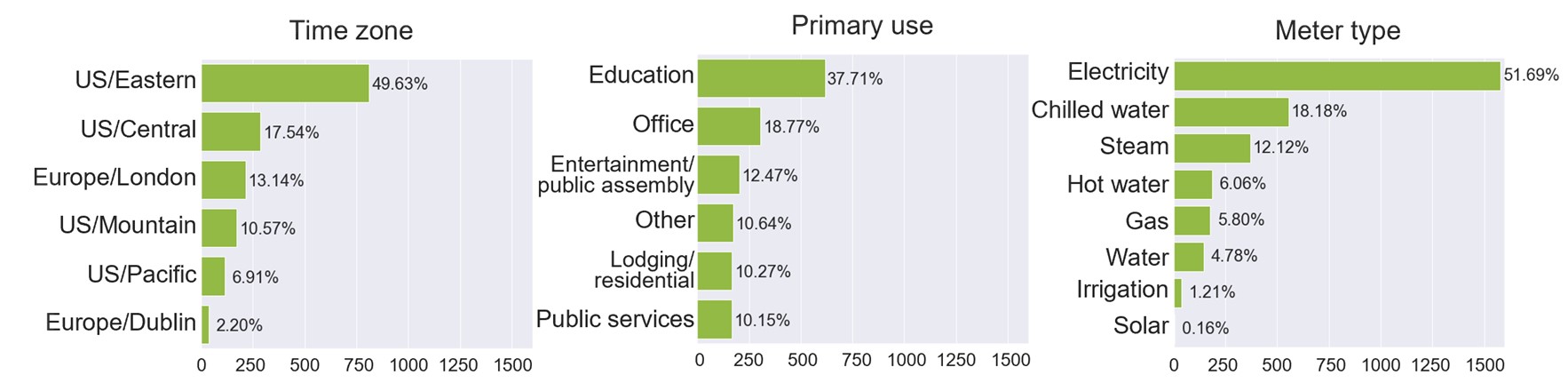}
\caption{Counts and categories of building-metadata features in BDG2 dataset \cite{miller2020building}.}
\label{fig:bdg_freq}
\end{center}
\end{figure*}

Each meter in the dataset comes with associated metadata, including geographic location, building type, and meter type (the categories and counts can be found in Figure \ref{fig:bdg_freq}). These metadata attributes are crucial for this study, as they guide the model in generating energy data that conforms to specified contextual factors. 
Meter types (e.g., electricity, hot water, and chilled water) and building types (e.g., education and office) represent energy usage types and behaviors and are therefore included as generation conditions.
The selection of these metadata attributes is informed by their demonstrated strong influence on energy usage, as found in previous studies \cite{miller2020ashrae, fu2022using}.
Although weather data is known to have a high impact on energy usage, we simplified the generative model by not directly using it as an input variable to facilitate training. 
Instead, we incorporated geographic location, represented by latitude and longitude, into the generative model to reflect the influence of different geographical areas.
The ultimate goal is to generate annual hourly data that corresponds to these metadata attributes. For instance, the model would produce year-round energy data for chilled water meters situated in office buildings at specific geographic locations.

\subsection{Data preprocessing}
The energy dataset undergoes a series of preprocessing steps to set the stage for the subsequent development of the deep generative model. These measures aim to optimize the quality of the data and include the removal of anomalous meter readings through data cleaning, standardizing the dataset to a common range, and splitting the dataset into train and test subsets. This not only ensures high-quality data but also readies the dataset for subsequent model training and testing.

\subsubsection{Data cleaning}
Data preprocessing serves as a crucial initial step before delving into modeling. Through initial exploratory analysis, we identify gaps and anomalies within the dataset that must be addressed to facilitate subsequent model development. 
Outliers are detected and removed using the Interquartile Range (IQR) method. Missing values are handled using forward and backward fill methods, where missing values are filled based on the next and previous observed values, respectively. Moreover, columns with a higher percentage of missing values (more than 5\%) are excluded from the dataset to preserve the integrity of the analysis. This approach ensures that anomalous and missing readings do not skew the results. After the cleaning process, the dataset comprises 1,828 meters.

\subsubsection{Data normalization}
\begin{figure}[!htb]
\begin{center}
\includegraphics[width=0.4\textwidth, trim= 0cm 0cm 0cm 0cm,clip]{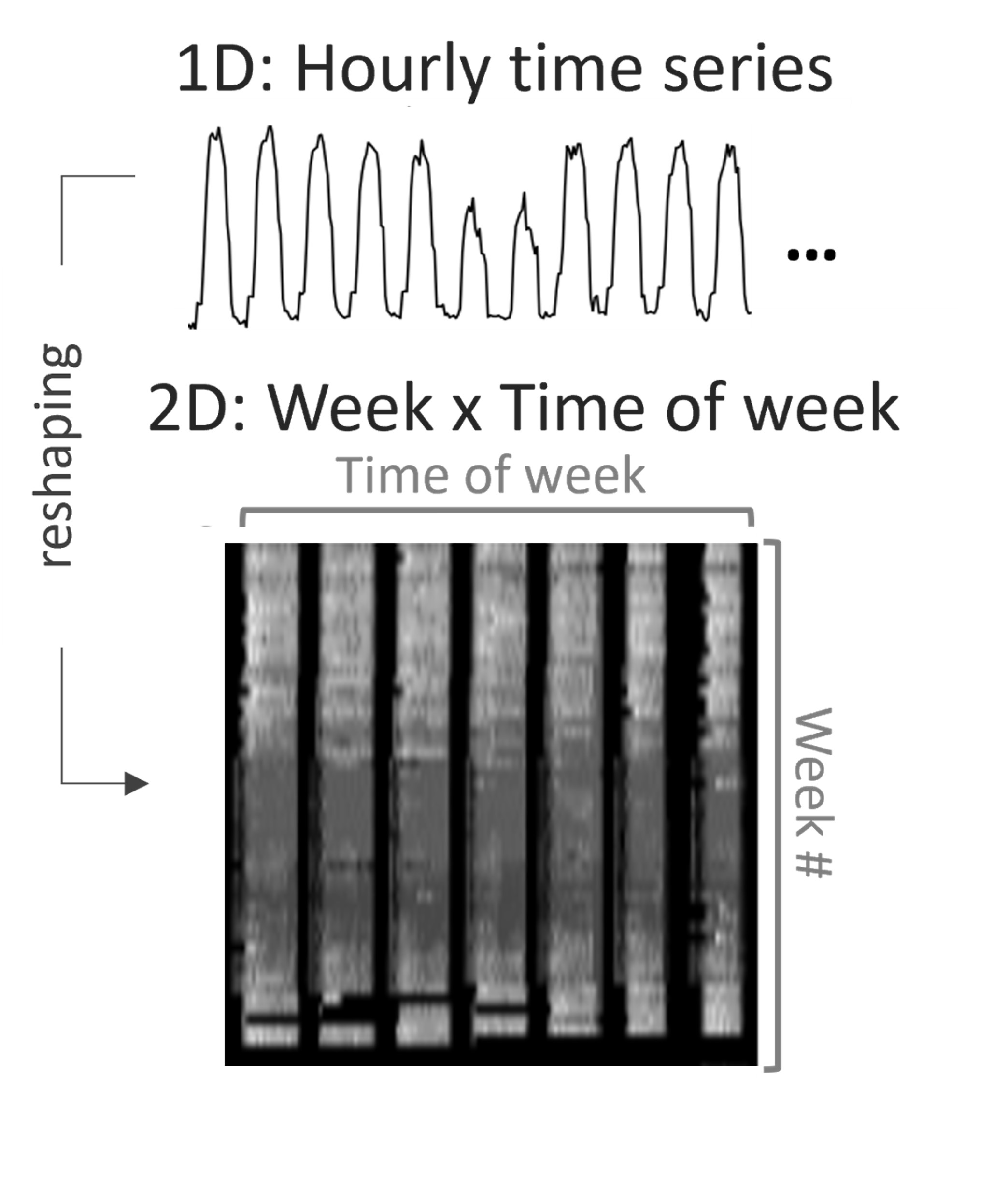}
\caption{Original 1D time series energy data were reshaped into 2D data to capture weekly usage patterns and then integrated into the image-based generative model.}
\label{fig:reshaping_illustration}
\end{center}
\end{figure}

Following data cleaning, normalization is conducted to standardize the features' range. Each data point is scaled down to a common range between -1 and 1. Min-max scaling is applied across the dataset to achieve this standardization. Normalizing the data ensures that no individual feature disproportionately influences the model training, making the optimization problem easier to solve. 
Additionally, certain features, such as meter types and building types, are encoded to make them usable within the model.
Furthermore, the data is reshaped to conform to the appropriate dimensions required for integration into the image-based generative model (as depicted in Figure \ref{fig:reshaping_illustration}). This reshaping enhances the model's capacity to capture the contextual aspects of the data.

\subsubsection{Data splitting}
The final preprocessing step involves partitioning the dataset into training and testing subsets.
75\% of the meter data is randomly selected to form the training dataset, and the remaining 25\% is used for testing.
This random partitioning aids in evaluating the model on unseen meter data, thus providing an unbiased performance evaluation.

\subsection{Modeling}

To enable the generation of high-fidelity synthetic energy data, this study implements and compares three state-of-the-art conditional generative models: Conditional Variational Autoencoder (CVAE), Conditional Generative Adversarial Network (CGAN), and a novel conditional diffusion model tailored for energy data synthesis. These models are conditioned on relevant metadata to capture intricate dependencies in energy consumption patterns. The following subsections provide an overview of each approach.

\subsubsection{Conditional Variational Autoencoder (CVAE)}

The Conditional Variational Autoencoder (CVAE) is an advancement of the traditional Variational Autoencoder (VAE), incorporating conditionality on specific variables to enhance the variational inference framework \cite{sohn2015learning}. In the context of synthetic energy data generation, this study exploits the capabilities of CVAEs to condition building metadata, thereby capturing the underlying dependencies in energy consumption patterns. The model comprises an encoder and a decoder neural network; the encoder takes the energy data and metadata as inputs and compresses them into a latent space. A sampling function introduces randomness to this latent representation, from which the decoder, conditioned on the metadata, attempts to reconstruct the energy data.

\begin{figure*}[!htb]
\begin{center}
\includegraphics[width=0.7\textwidth, trim= 0cm 0cm 0cm 0cm,clip]{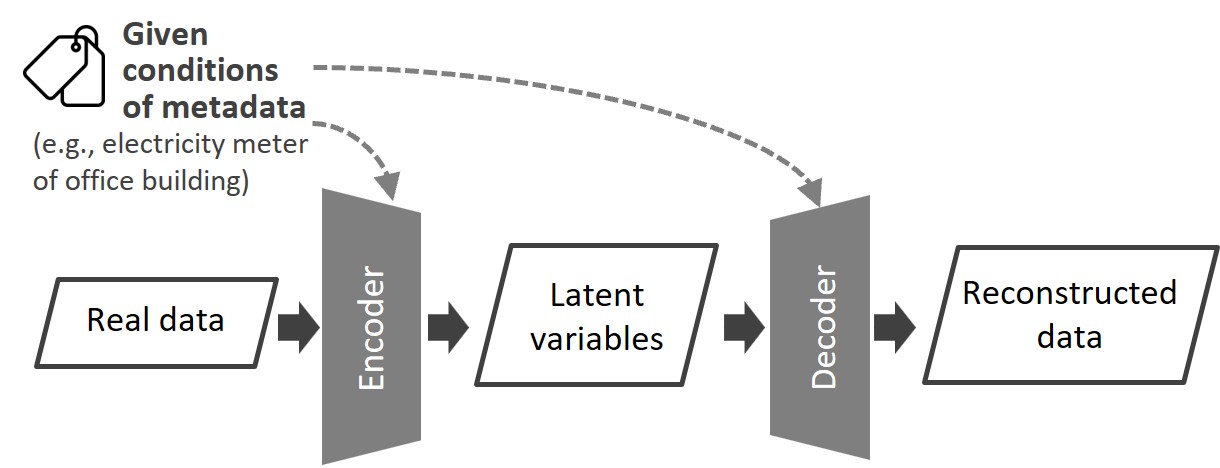}
\caption{Illustration of the CVAE model coupled with metadata. The trained decoder takes latent variables and metadata as input, enabling the generation of conditional energy data.}
\label{fig:CVAE_illustration}
\end{center}
\end{figure*}

The incorporation of conditional variables like building type and meter type distinguishes CVAEs and guides the data generation process \cite{bao2017cvae}. This tailored conditioning allows the generation of high-fidelity synthetic energy data that respects the intricate relationships present in real data while still capturing inherent stochasticity. Nonetheless, a limitation lies in the restricted expressivity of the prescribed prior distribution on the latent space, which may not fully encapsulate the diverse range of real energy data. 
Figure \ref{fig:CVAE_illustration}) illustrates the flow from real data through the encoder to latent variables and then through the decoder to reconstructed data, conditioned on metadata. This decoder, equipped with latent variables and metadata, can then be employed for the generation of synthetic energy data.

\subsubsection{Conditional Generative Adversarial Networks (CGANs)}

CGANs extend the Generative Adversarial Network (GAN) framework by incorporating auxiliary information, often referred to as metadata or labels, for conditioning both the generator and discriminator \cite{mirza2014conditional}. In this research, the CGAN model is conditioned explicitly on building metadata such as meters and building types to generate synthetic energy consumption data. This allows the network to produce data that not only resembles real-world energy usage patterns but is also coherent with the given building context.

The CGAN framework comprises a generator model, which aims to produce synthetic energy data, and a discriminator model, tasked with differentiating between real and synthetic data. These two components are trained iteratively in an adversarial setting until both networks converge, and the generated data becomes indistinguishable from real-world data.

\begin{figure*}[!htb]
\begin{center}
\includegraphics[width=0.7\textwidth, trim= 0cm 0cm 0cm 0cm,clip]{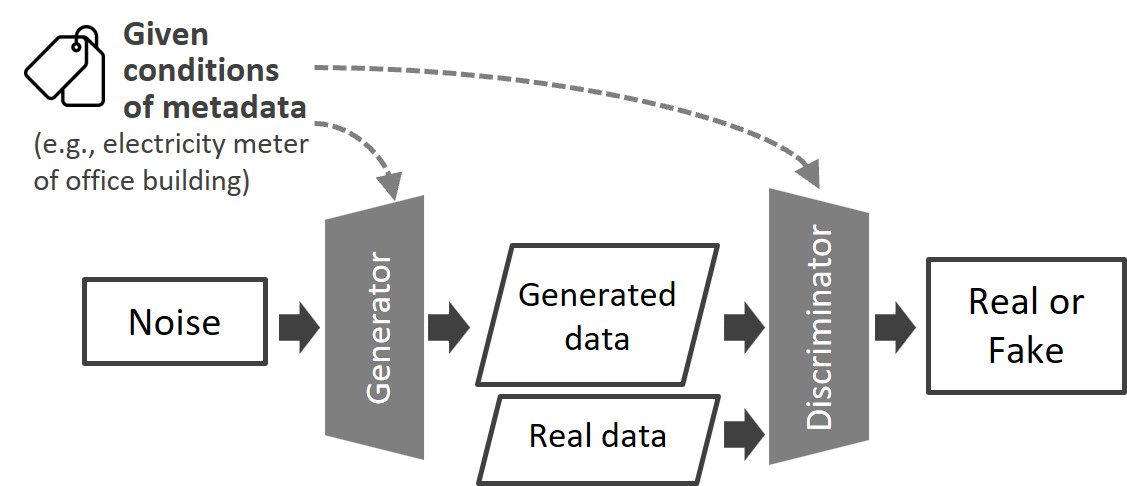}
\caption{Illustration of the CGAN model combined with metadata. The discriminator aids the generator in producing realistic samples, while the trained generator can generate synthetic data conditioned on the provided metadata.}
\label{fig:CGAN_illustration}
\end{center}
\end{figure*}

One of the key advantages of CGANs is their ability to produce high-quality samples, a feature attributed to the adversarial training approach \cite{odena2017conditional}. By utilizing adversarial loss instead of relying on restrictive probabilistic models, CGANs are adept at capturing complex data distributions. However, they are also known for their training challenges, including mode collapse — a phenomenon where the generator can only produce limited types of outputs  \cite{che2016mode, kushwaha2020study}. This issue arises due to the volatile competitive balance between the generator and discriminator within their framework. As such, meticulous architecture design and training methodology are essential for the successful application of CGANs in the energy domain. 
Figure \ref{fig:CGAN_illustration} depicts the CGAN architecture used in this study, highlighting the interaction between the generator and discriminator, with the incorporation of metadata influencing the generation process. The generator is guided by metadata to synthesize data that is then evaluated by the discriminator for authenticity.

\subsubsection{Conditional diffusion model}

Diffusion models have emerged as a promising generative modeling paradigm, reversing a noise injection process to transform pure noise into samples from the data distribution \cite{ho2020denoising}. Unlike traditional generative approaches like GANs and VAEs, which sometimes suffer from training instability and hyperparameter sensitivity, diffusion models offer advantages in training stability, ease of hyperparameter tuning, and high sample quality \cite{dhariwal2021diffusion}. 

\begin{figure*}[!htb]
\begin{center}
\includegraphics[width=0.85\textwidth, trim= 0cm 0cm 0cm 0cm,clip]{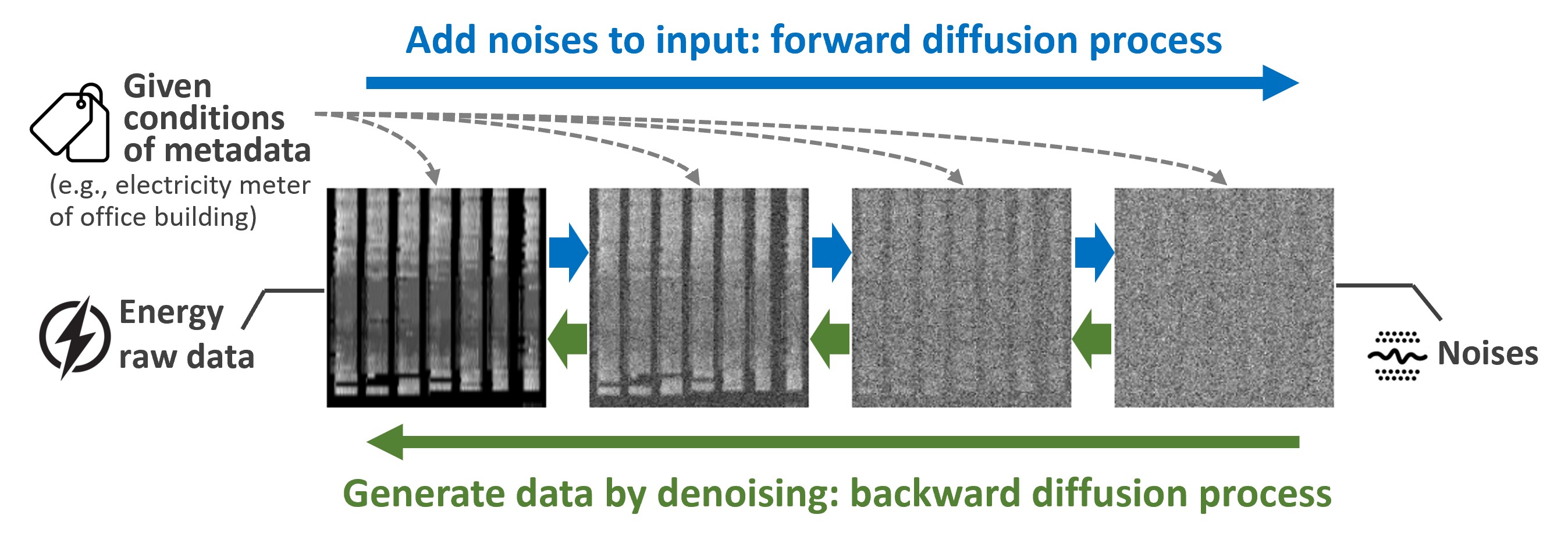}
\caption{Illustration of the proposed meta-driven conditional diffusion model. The backward diffusion process generates data by progressive denoising, resulting in realistic energy data conditioned on metadata attributes.}
\label{fig:diffusion_illustration}
\end{center}
\end{figure*}

In this context, we introduce an advanced conditional diffusion Mmodel tailored for synthesizing high-quality, long-term energy data. This model employs a context-dependent U-Net architecture to incorporate metadata about energy consumption patterns. Further enhancing its capabilities, a time-embedding layer is integrated to capture the temporal dependencies prevalent in energy data. Our conditional diffusion model starts with a noise base and progressively denoises it into realistic energy data, which are aligned with encoded metadata attributes. This dual capability of maintaining data integrity while factoring in rich contextual and temporal information makes it particularly well-suited for handling complex but structured energy datasets.
Figure \ref{fig:diffusion_illustration} illustrates the forward diffusion process where noise is incrementally added to the energy raw data, transforming it into pure noise. Conversely, the backward diffusion process involves the generation of data by denoising, starting from noise and progressively restoring the data to generate realistic energy load profiles conditioned by metadata attributes.

\subsubsection{Hyperparameters setting}
This research is built upon the foundation of three key models: CVAE, CGAN, and conditional diffusion model. Each of these models is tailored with distinct hyperparameter configurations, carefully chosen to strike a balance between performance and computational demands. The architecture of the CVAE model incorporates convolutional layers in the encoder and transposed convolutional layers in the decoder. For the encoder's Conv2D layers, filters of sizes $16$, $32$, and $64$ with a kernel size of $3 \times 3$ and ReLU activation functions are utilized. A dense layer with $64$ dimensions is employed before the latent space with 512 dimensions. Correspondingly, the decoder contains Conv2DTranspose layers with $64$, $32$, and $16$ filters with the same kernel size and activation functions. The optimization algorithm used is Adam, with Mean Squared Error (MSE) used for the reconstruction loss and KL divergence loss utilized for optimizing the encoded latent space.

The architecture of the CGAN model consists of convolutional layers in the discriminator and transposed convolutional layers in the generator. For the discriminator's Conv2D layers, filters of sizes \(32\), \(64\), and \(128\) with a kernel size of \(3 \times 3\) and leaky ReLU activation functions are used, while the generator contains Conv2DTranspose layers with \(128\), \(64\), and \(32\) filters. The generator is initiated with a noise vector of 100 dimensions. The optimization algorithm applied is Adam with a learning rate of \(0.0001\). The loss function employed for training is binary cross-entropy. Both the generator and the discriminator are conditioned on building metadata, which aligns with the dimensions and context of the test set. 

The architecture of the Conditional Diffusion Model is designed around a U-Net structure. The model employs a time-embedding layer for capturing temporal patterns and a context vector for incorporating metadata. The diffusion process operates over 500 timesteps, each introducing noise following a predefined schedule, starting with an initial noise scale (\(\beta_1\)) of \(1 \times 10^{-4}\) and gradually reaching a final noise scale (\(\beta_2\)) of 0.02. The hidden layer used in U-Net has 64 features that are optimized to capture the essence of energy data. A context vector, aligned with the metadata, has a dimensionality dictated by the metadata's size. For optimization, the Adam algorithm is employed, targeting to reduce loss function in MSE with a learning rate of \(1 \times 10^{-3}\) and a batch size of 10. This hyperparameter configuration is selected based on experiments aimed at achieving an equilibrium between computational efficiency and modeling accuracy, tailored to real-world energy data. 

In this research, early stopping was intentionally not applied during the training of the generative model. One primary reason for this decision is the unreliable nature of loss values as performance indicators in the context of generative models, as highlighted by \cite{borji2019pros}. On the other hand, exclusive reliance on qualitative human evaluations for determining sample quality can yield incomplete and potentially misleading conclusions about the model's performance \cite{theis2015note}. Therefore, the number of training iterations was determined by achieving a stable level of training loss, while other metrics, such as KL divergence and FID score, were also monitored to evaluate quality and diversity. This approach ensures a more comprehensive assessment of model performance.

\subsubsection{Evaluation metrics}
Evaluating synthetic data, especially in the context of energy datasets, requires a meticulous approach that is able to assess both the distinctiveness and subtleties of the generated data. In order to ensure that the synthetic data not only adheres closely to real data distributions but also maintains temporal coherence and variability, we enlist a suite of evaluation metrics, each of which addresses a particular aspect of synthetic data quality.

To scrutinize the \textit{generative variability}, we employ the Fr\'echet Inception Distance (FID), a widely recognized metric in the domain of generative modeling. FID measures the distance between the distributions of generated and real data in the feature space, which were calculated using the pre-trained inception-v3 model used for image classification \cite{heusel2017gans}. It can effectively evaluate how well the model captures the data diversity present in the genuine dataset. Lower FID scores correlate with superior diversity of generated samples, implying a closer match between synthetic and real data distributions.

For evaluating \textit{time-series prediction} quality, Root Mean Squared Error (RMSE) and the Coefficient of Determination (\(R^2\)) are leveraged. RMSE provides a measure of the average deviation of predicted values from the observed values, offering a straightforward, interpretable metric for prediction accuracy. \(R^2\), on the other hand, gives insight into the predictive power of the model by indicating the proportion of variance in the dependent variable that is predictable from the independent variables. 
It is worth noting that while RMSE and \(R^2\) may not be ideal for evaluating generative tasks, they do offer valuable insights into how contextual information or conditions can be effectively translated into target time series from a predictive view.

To ascertain the \textit{distribution similarity} between synthetic and real data, we utilize the Kullback-Leibler (KL) divergence, which quantifies how one probability distribution diverges from a second, expected probability distribution \cite{kullback1951information}. Minimizing KL divergence ensures that the generated synthetic data adheres as closely as possible to the original data in a probabilistic sense, ensuring that the synthetic data is not only visually but also statistically coherent.

These metrics, FID, RMSE, $R^2$, and KL divergence, collectively forge a rigorous evaluation framework for synthetic data, ensuring that the generated data is critically assessed for both visual and statistical fidelity across various dimensions, providing a holistic view of model performance in synthetic data generation.

\subsection{Experiment design}

The experimental design for this research is meticulously tailored to address the generative task of synthesizing long-term annual energy load data. Detailed specifications of the experimental setup are provided in Table \ref{tab:exp_design}. The input for the experiment consists of metadata from 1,828 meters, including various parameters such as year, location (latitude and longitude), types of meters (e.g., electricity, chilled water, gas, hot water, and steam), and the associated building types. This metadata serves to generate synthetic energy data that closely align with real-world conditions.

\begin{table}[!htb]
\centering
\resizebox{0.5\textwidth}{!}{%
\begin{tabular}{ll}
\hline
 & \textbf{Setting of Experiments} \\ \hline
Input & \begin{tabular}[c]{@{}l@{}}Meta data of 1828 meters\\   - Year: 2016, 2017\\   - Location: Latitude and longitude\\   - Meter types: electricity, chilled water, gas, \\ hotwater, steam\\   - Building types: Office, Education, Public \\ services, Entertainment, Lodging/residential\end{tabular} \\
\\
Output & Hourly data of 1828 annual time series \\
\\
Data split & 75\% for training; 25\% meters for testing \\
\\
Models & CGAN, CVAE, and conditional diffusion model \\
\\
Metrics & \begin{tabular}[c]{@{}l@{}}- Generative diversity: FID\\ - Distribution similarity: KL divergence\\ - Time-series prediction: RMSE and $R^2$\end{tabular} \\ \hline
\end{tabular}
}
\caption{Details of modeling and experiment settings of the study in synthetic data generation.}
\label{tab:exp_design}
\end{table}



\section{Results}
This section presents the key findings from the comparative evaluation of the generative models. It details the quantitative performance across metrics and sample visualizations, analyzing the capabilities of the CVAE, CGAN, and proposed Diffusion model in synthesizing diverse, accurate long-term building energy data.

\subsection{Overview of generation performance across models}

The primary objective of this research is to develop an efficient generative model for synthesizing high-quality energy data. In assessing the quality of the synthetic data generated, the evaluation metrics offer a holistic understanding of the performance of our proposed conditional diffusion model, especially when compared to other baseline generative models. Detailed metric values for each model are presented in Table \ref{tab:step3_metrics}, and all scores/errors' ranges were calculated through 30 iterations of data generation, each with different random seeds.

\begin{table*}[!htb]
\centering
\resizebox{1.0\textwidth}{!}{%
\begin{tabular}{lllllllllllll}
\hline
\textbf{Model} & \textbf{} & \multicolumn{3}{l}{\textbf{FID score}} & \multicolumn{3}{l}{\textbf{KL divergence}} & \multicolumn{3}{l}{\textbf{\begin{tabular}[c]{@{}l@{}} $R^2$ \end{tabular}}} & \multicolumn{2}{l}{\textbf{RMSE}} \\ \hline
CVAE &  & 946.7 & $\pm$10.1 &  & 0.45 & $\pm$0.0038 &  & 0.35 & $\pm$0.0071 &  & 0.26 & $\pm$0.0017 \\
CGAN &  & 809.3 & $\pm$5.1 &  & 0.88 & $\pm$0.0096 &  & 0.33 & $\pm$0.0023 &  & 0.27 & $\pm$0.00054 \\
Diffusion model &  & \textbf{517.3} & \textbf{$\pm$2.1} &  & \textbf{0.40} & \textbf{$\pm$0.0013} &  & \textbf{0.43} & \textbf{$\pm$0.0021} &  & \textbf{0.25} & \textbf{$\pm$0.00052} \\ \hline
\end{tabular}
}
\caption{Summary of metrics across generative models with 30 iterations for calculating the score/error interval within one standard deviation. The best scores are highlighted in bold font (lower values are better for all metrics except for $R^2$).}
\label{tab:step3_metrics}
\end{table*}

As shown in Table \ref{tab:step3_metrics}, KL divergence values indicate that the proposed diffusion model performs better in capturing the statistical characteristics of the original data. A lower KL divergence of \(0.40 \pm 0.0013\) signifies that the generated data closely mirrors the true data distribution. On the other hand, FID scores are employed to measure the variability of the generated datasets. The proposed model achieved a FID score of \(517.3 \pm 2.1\), which is considerably lower than that of competing models. These lower FID scores signify higher diversity in the generated data, confirming the model's ability to produce data with varying energy load patterns.
With regard to time-series prediction accuracy, RMSE and $R^2$ serve as effective indicators. The diffusion model yields an RMSE of \(0.25 \pm 0.00052\), demonstrating a better fit to the observed data when contrasted with other models under consideration. Concurrently, the $R^2$ aids in quantifying the linear relationship between the generated and actual data sets. A higher $R^2$ of \(0.43 \pm 0.0021\) further confirms that the model is more capable of preserving structural relationships within the data compared to other models. Nonetheless, in the context of energy forecasting, a correlation of 0.43 is not entirely satisfactory; however, given that the model is generative and leverages metadata for predictions, these conventional time-series metrics are not the sole indicators for evaluating performance. Collectively, these metrics, together with FID and KL divergence, suggest that the proposed diffusion model is more effective than other generative models in synthesizing long-term annual load data.

\begin{figure*}[!htb]
\begin{center}
\includegraphics[width=0.8\textwidth, trim= 0cm 0cm 0cm 0cm,clip]{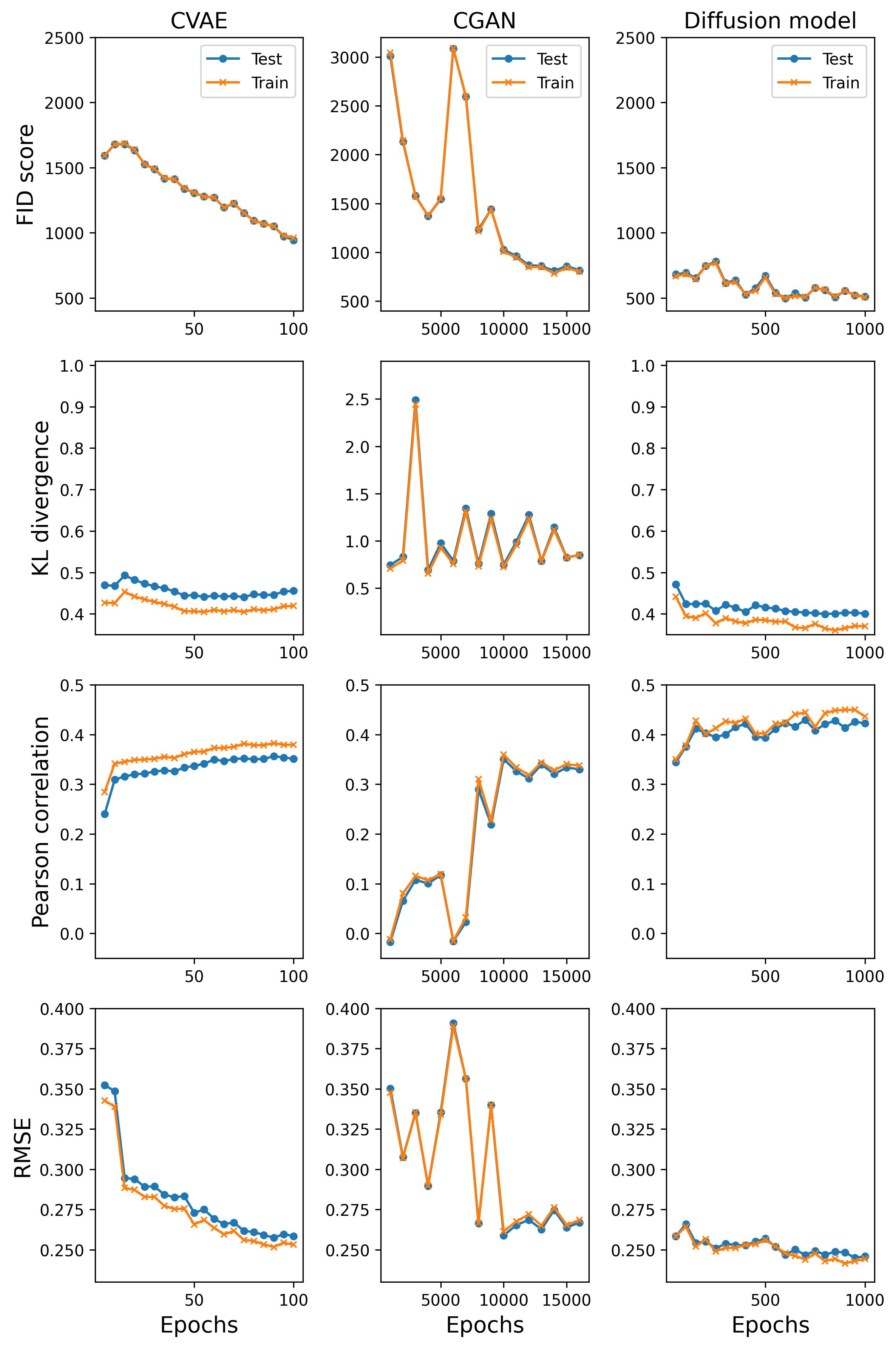}
\caption{Line plots of evaluation metrics on training and test data across training epochs for various models.}
\label{fig:epochs_plots}
\end{center}
\end{figure*}

The evolution of model performance during training epochs is captured in Figure \ref{fig:epochs_plots}. 
The plots show that the Conditional Variational Autoencoder (CVAE) showcases stable and gradually improving performance throughout its training process with 100 training epochs. This stability and gradual improvement may be attributed to the Gaussian assumption of latent space distribution in the CVAE architecture, which tends to learn the underlying data distribution effectively and generally.
In contrast, the Conditional Generative Adversarial Network (CGAN) exhibits larger fluctuations in its performance, as reflected by its 15,000 training epochs. These fluctuations can be attributed to the inherent challenges associated with GAN architectures, including issues like mode collapse, which can hinder the stability of the training process. The CGAN's performance may show periods of rapid improvement followed by periods of stagnation or even degradation, contributing to the observed variability in its performance. 
Meanwhile, the diffusion model, with 1,000 training epochs, not only demonstrates a stable performance but also consistently outperforms both the CVAE and CGAN in terms of the metrics considered. This stability in performance can be attributed to the unique properties of the diffusion model, which leverages diffusion processes to generate data samples. These processes allow for more stable and consistent learning, resulting in superior performance compared to the other generative models under consideration.

It is noteworthy that since the losses implemented in the generative models are diverse and not based on these evaluation metrics, early stopping was not made in training. The number of training epochs was determined by a comprehensive consideration of all metrics, ensuring the termination of training once a stable convergence was observed across the metrics and balancing model stability and performance.

\subsection{Analysis of breakdown by meter and building types}
\begin{figure*}[!htb]
\begin{center}
\includegraphics[width=1.0\textwidth, trim= 0cm 0cm 0cm 0cm,clip]{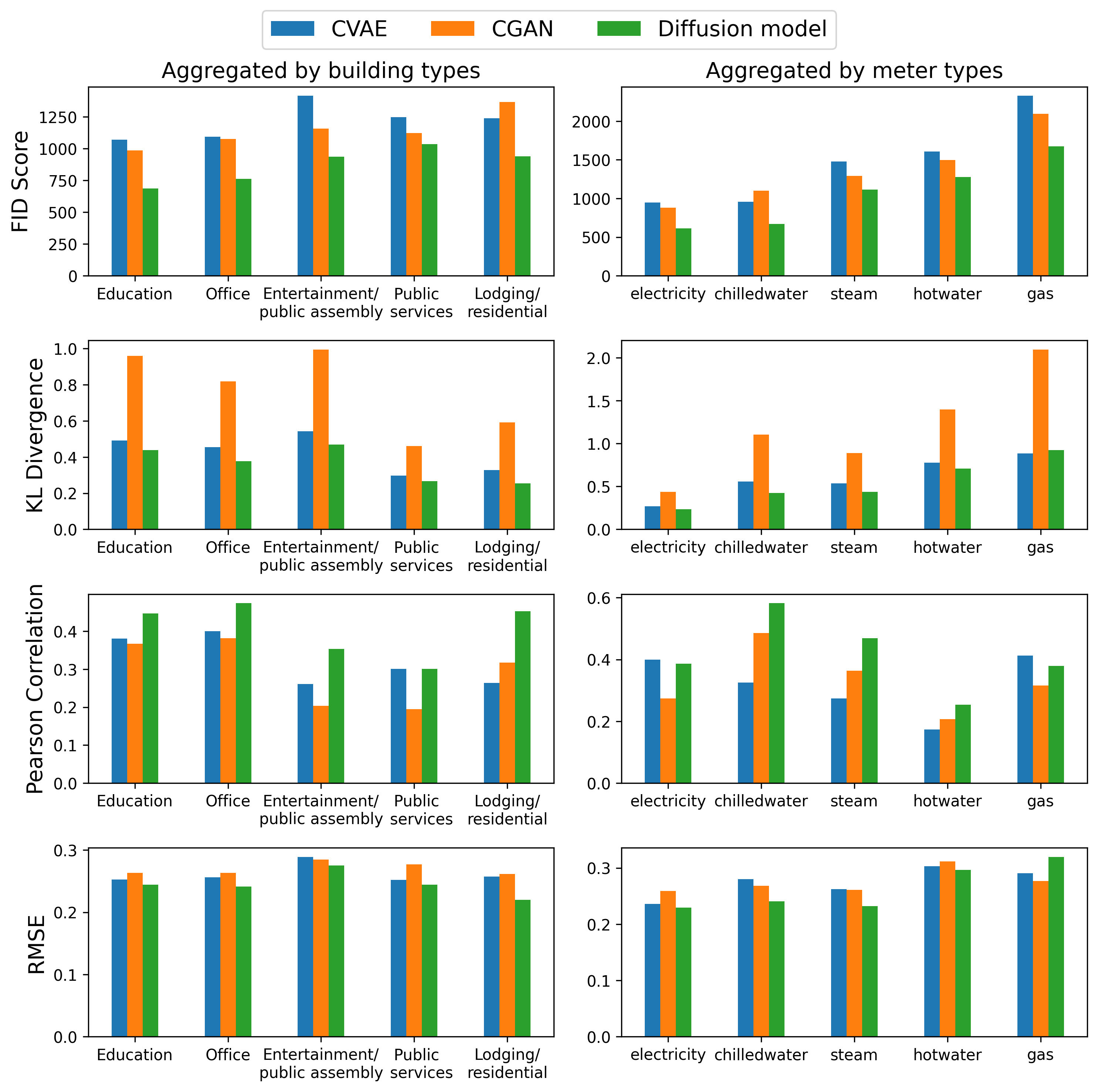}
\caption{Comparative analysis of different metrics across models.}
\label{fig:metrics_comparison}
\end{center}
\end{figure*}

The results in Figure \ref{fig:metrics_comparison} display a thorough analysis of various generative models categorized by building types and meter types. Starting with the FID score and KL divergence, it is evident that, within diverse building categories, the diffusion model consistently surpasses both the CVAE and CGAN models. This suggests that the proposed diffusion model effectively captures the diversity and distribution likeness of the energy data. As for the CVAE and CGAN, which are secondary in performance, an intriguing observation can be made from their comparative performance. While CGAN exhibits higher diversity with a lower FID score, its fidelity, as indicated by KL divergence, is noticeably inferior. This discrepancy mainly stems from the CVAE's tendency to provide more generalized and consistent predictions, whereas CGAN, due to its adversarial generation framework, produces more diverse results at the expense of fidelity.

Diving into the time-series metrics, both $R^2$ and RMSE are crucial in gauging the predictive accuracy of the generated datasets. For most building types, the diffusion model once again stands out, exhibiting higher $R^2$ values, indicating its enhanced ability to retain the intrinsic temporal patterns in the energy datasets. The dominance of this model is further emphasized when examining weather-dependent meter types, as it adeptly captures patterns for chilled water and hot water with heightened fidelity. On the other hand, the CVAE model struggles more when generating such weather-dependent meters, as it tends to produce regular, electricity-like meter data with limited variability.

Additionally, it's worth noting that the most prevalent building types, education and office, exhibit the best predictive quality and perform well across various metrics. Regarding meter types, electricity meters, which have a relatively lower correlation with weather, show better performance in metrics. Other weather-dependent meters, including chilled water, steam, hot water, and gas, appear more challenging in the generation tasks.

\subsection{Zoom in of generative results and quality analysis}
\begin{figure*}[!htbp]
    \centering
    
    \begin{subfigure}[b]{0.85\textwidth}
        \caption{Real data and generated data of \textbf{Electricity meters} in \textbf{Office buildings}}
        \includegraphics[width=\textwidth]{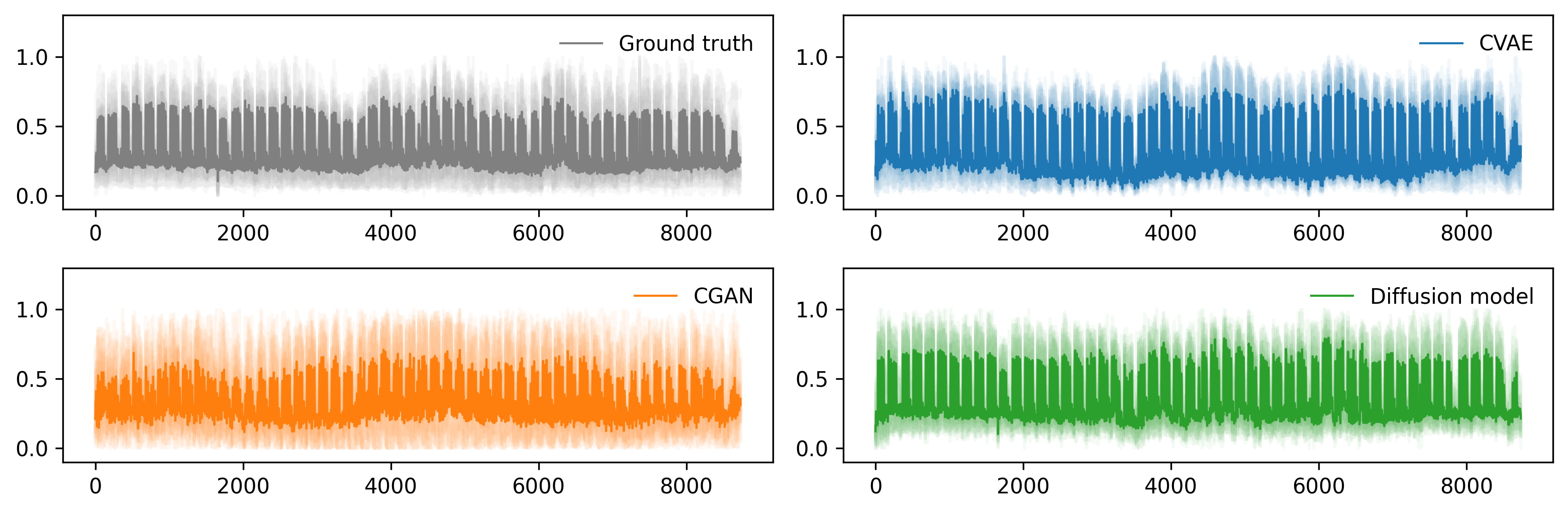}
        \label{fig:edu_elec_annual}
    \end{subfigure}
    
    \vspace{-2.0em}  
    
    \begin{subfigure}[b]{0.85\textwidth}
        \caption{Real data and generated data of \textbf{Electricity meters} in \textbf{Education buildings}}
        \includegraphics[width=\textwidth]{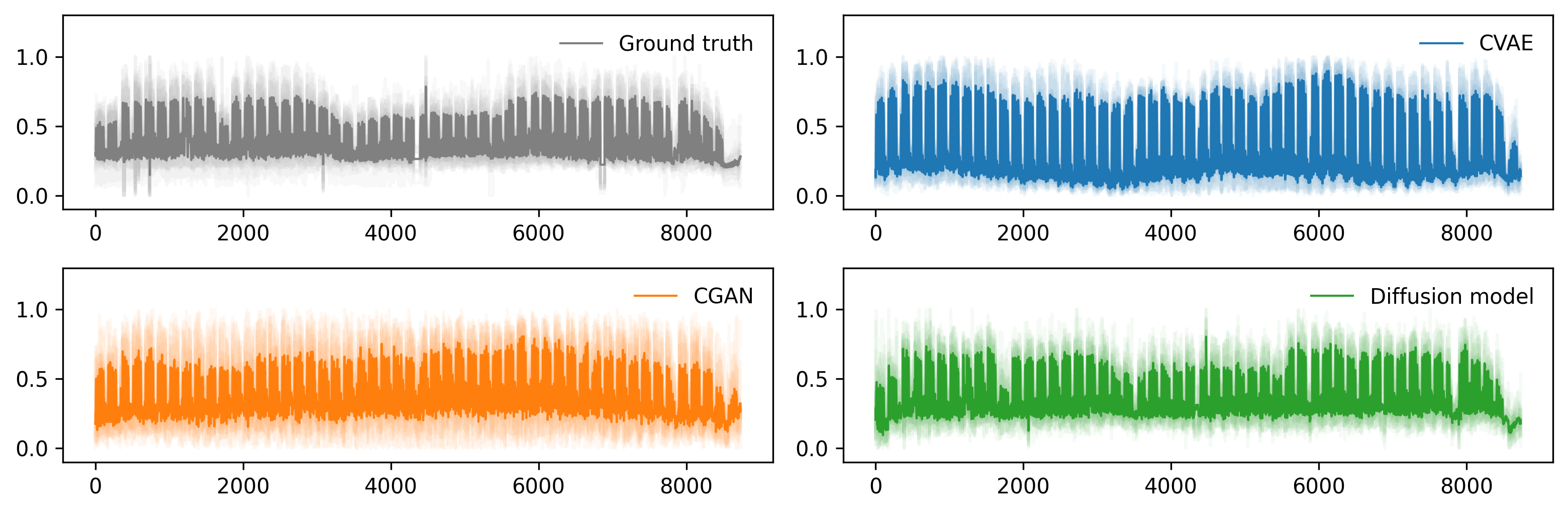}
        \label{fig:edu_chilledwater_annual}
    \end{subfigure}
    
    \vspace{-2.0em}  
    
    \begin{subfigure}[b]{0.85\textwidth}
        \caption{Real data and generated data of \textbf{Chilled water meters} in \textbf{Education buildings}}
        \includegraphics[width=\textwidth]{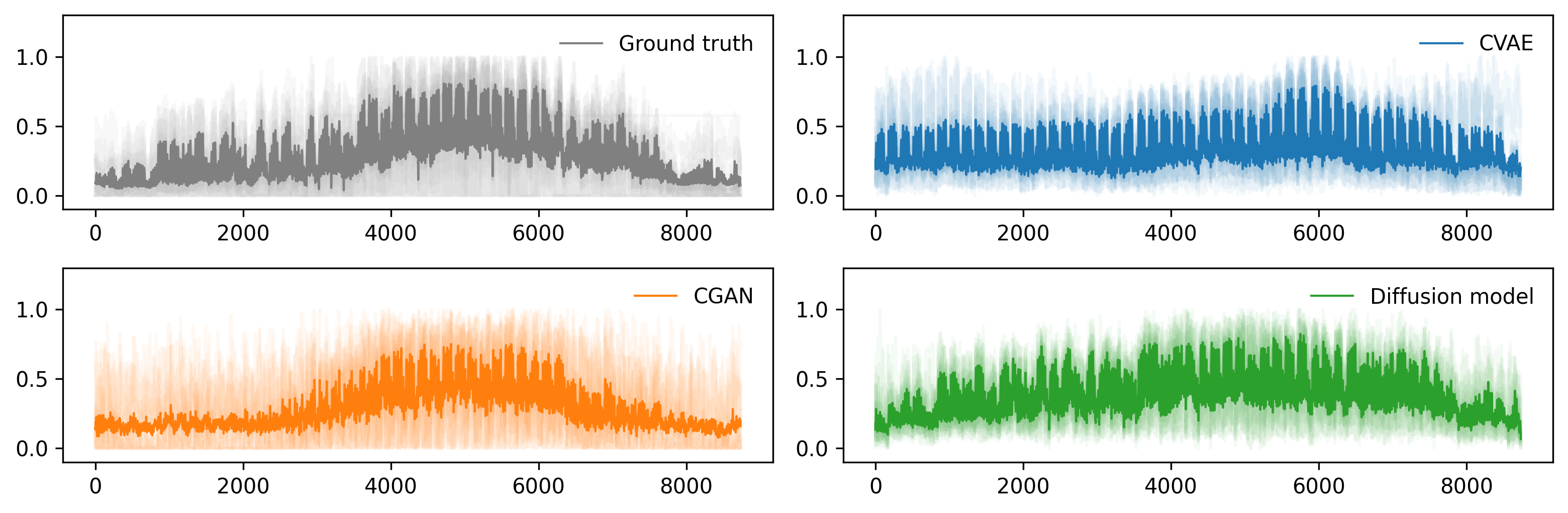}
        \label{fig:office_hotwater_annual}
    \end{subfigure}
    
    \vspace{-2.0em}  
    
    \begin{subfigure}[b]{0.85\textwidth}
        \caption{Real data and generated data of \textbf{Steam meters} in \textbf{Education buildings}}
        \includegraphics[width=\textwidth]{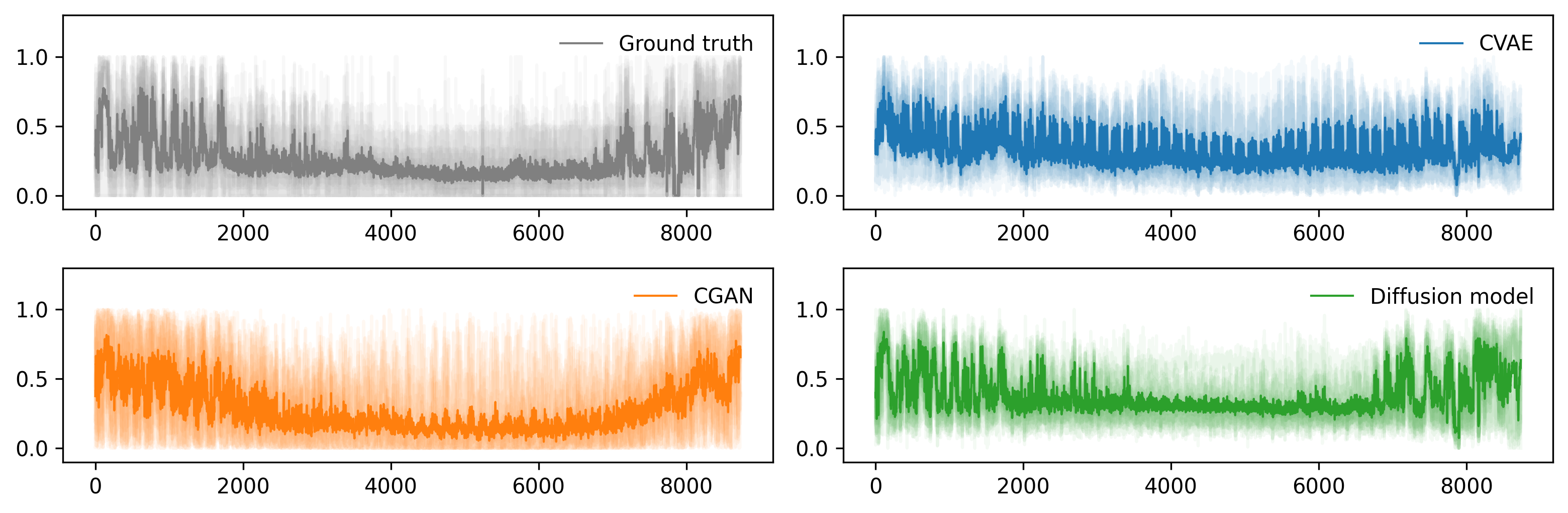}
        \label{fig:office_hotwater_annual}
    \end{subfigure}
    
    \caption{Comparison of ground truth and generated time series across models for different meter and building types with added transparency (hourly data visualized over a year).}
    \label{fig:clustered_ts_plots_annual}
\end{figure*}

\begin{figure*}[!htbp]
    \centering
    
    \begin{subfigure}[b]{0.85\textwidth}
        \caption{Real data and generated data of \textbf{Electricity meters} in \textbf{Office buildings}}
        \includegraphics[width=\textwidth]{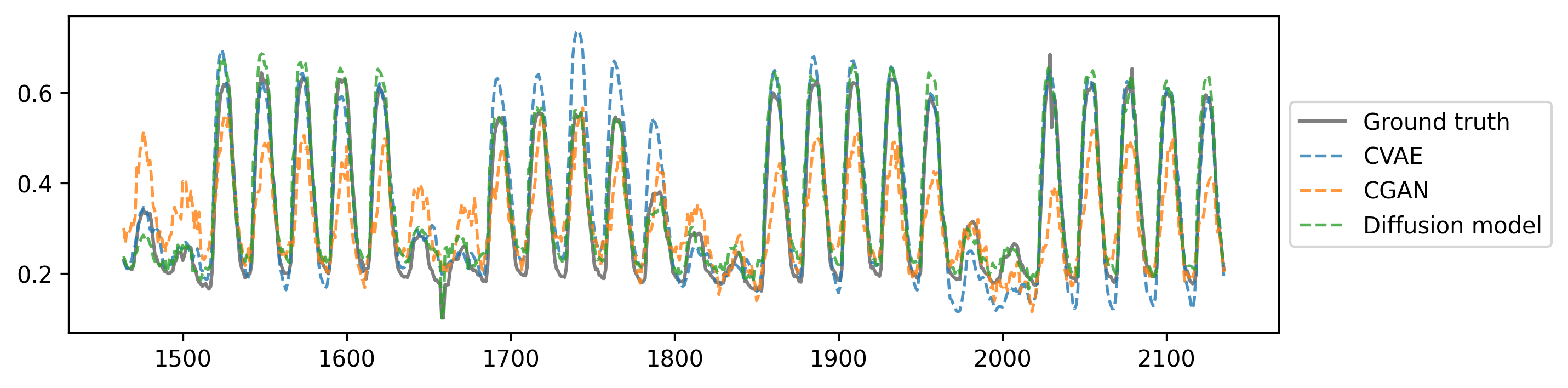}
        \label{fig:edu_elec}
    \end{subfigure}
    
    \vspace{-1.5em}  
    
    \begin{subfigure}[b]{0.85\textwidth}
        \caption{Real data and generated data of \textbf{Electricity meters} in \textbf{Education buildings}}
        \includegraphics[width=\textwidth]{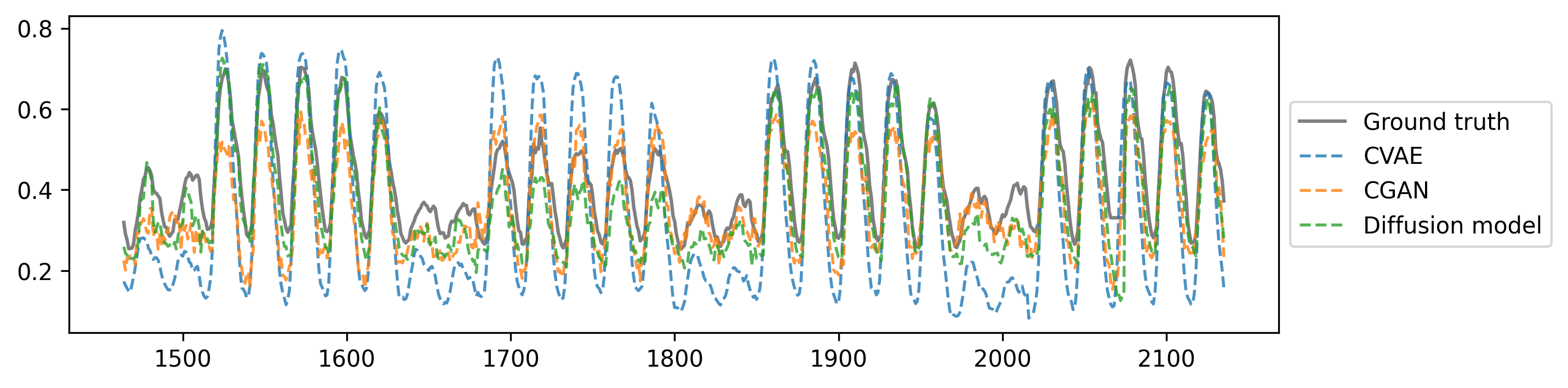}
        \label{fig:edu_chilledwater}
    \end{subfigure}
    
    \vspace{-1.5em}  
    
    \begin{subfigure}[b]{0.85\textwidth}
        \caption{Real data and generated data of \textbf{Chilled water meters} in \textbf{Education buildings}}
        \includegraphics[width=\textwidth]{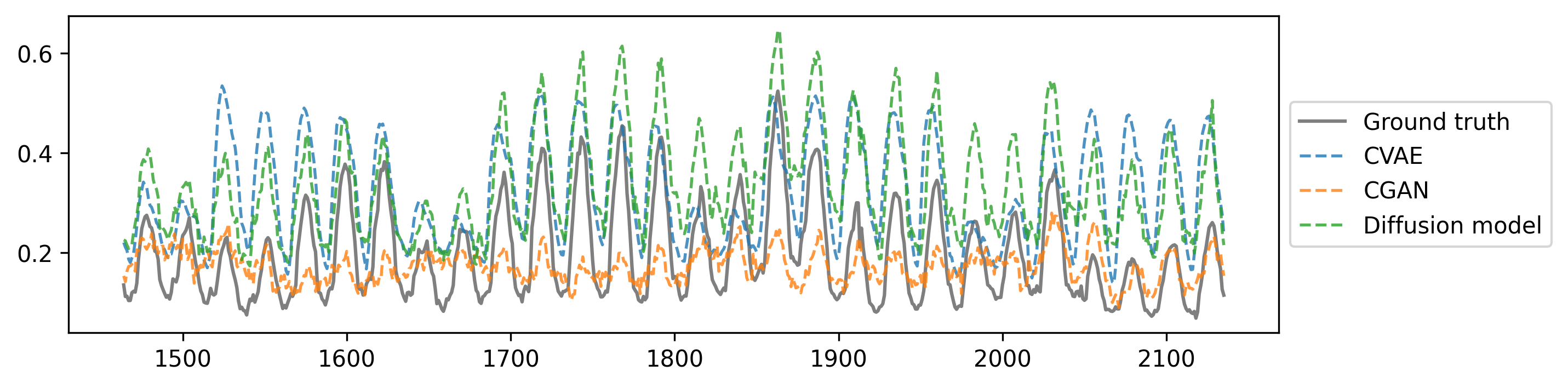}
        \label{fig:office_hotwater}
    \end{subfigure}
    
    \vspace{-1.5em}  
    
    \begin{subfigure}[b]{0.85\textwidth}
        \caption{Real data and generated data of \textbf{Steam meters} in \textbf{Education buildings}}
        \includegraphics[width=\textwidth]{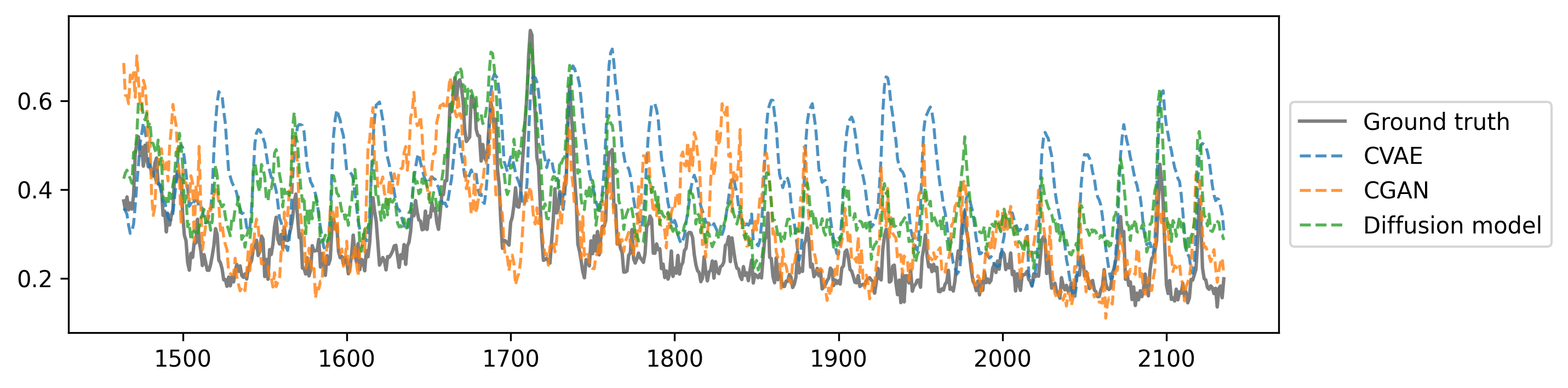}
        \label{fig:office_hotwater}
    \end{subfigure}
    
    \caption{Comparison of ground truth and generated time series across models for different meter and building types (hourly averaged data visualized over a month).}
    \label{fig:results_comp_plots}
\end{figure*}

\begin{figure*}[!htbp]
    \centering
    
    \begin{subfigure}[b]{0.85\textwidth}
        \caption{Real data and generated data of \textbf{Electricity meters} in \textbf{Office buildings}}
        \includegraphics[width=\textwidth]{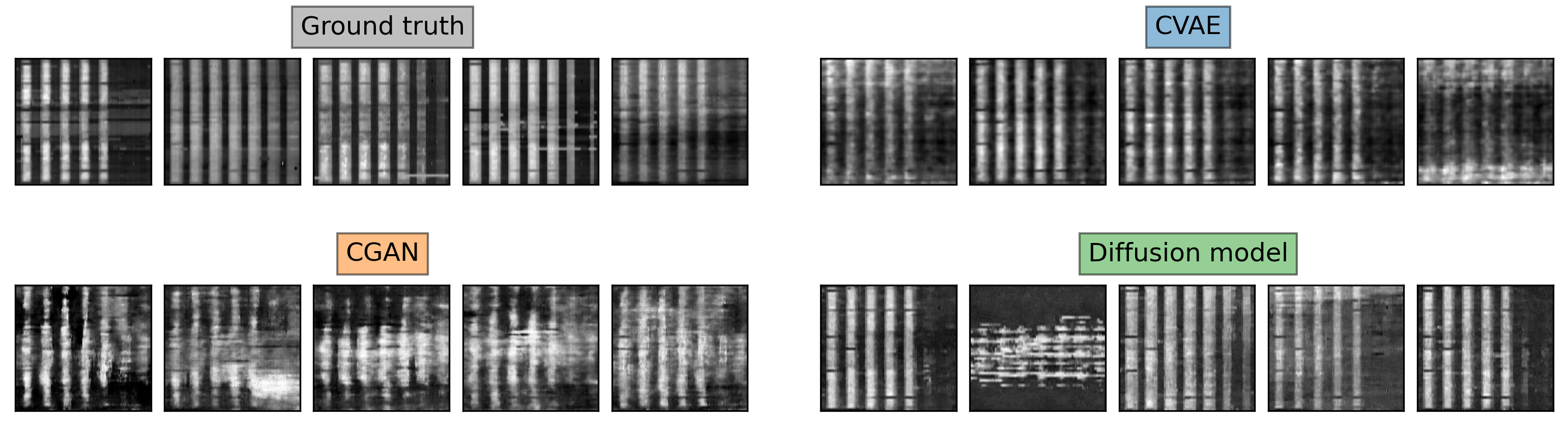}
        \label{fig:edu_elec}
    \end{subfigure}
    
    \vspace{-1.5em}  
    
    \begin{subfigure}[b]{0.85\textwidth}
        \caption{Real data and generated data of \textbf{Electricity meters} in \textbf{Education buildings}}
        \includegraphics[width=\textwidth]{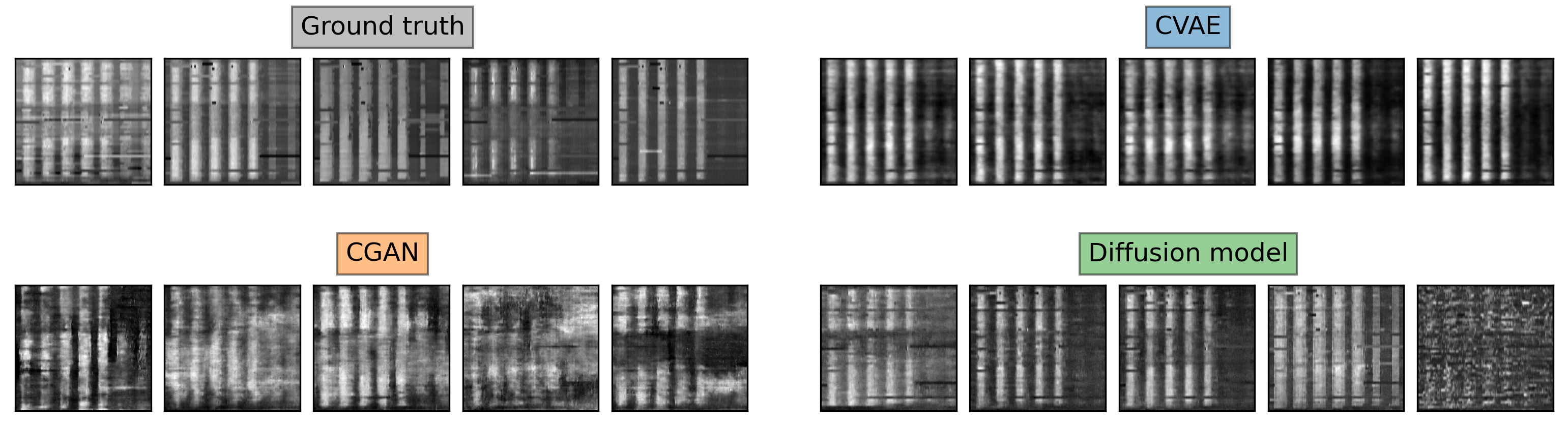}
        \label{fig:edu_chilledwater}
    \end{subfigure}
    
    \vspace{-1.5em}  
    
    \begin{subfigure}[b]{0.85\textwidth}
        \caption{Real data and generated data of \textbf{Steam meters} in \textbf{Education buildings}}
        \includegraphics[width=\textwidth]{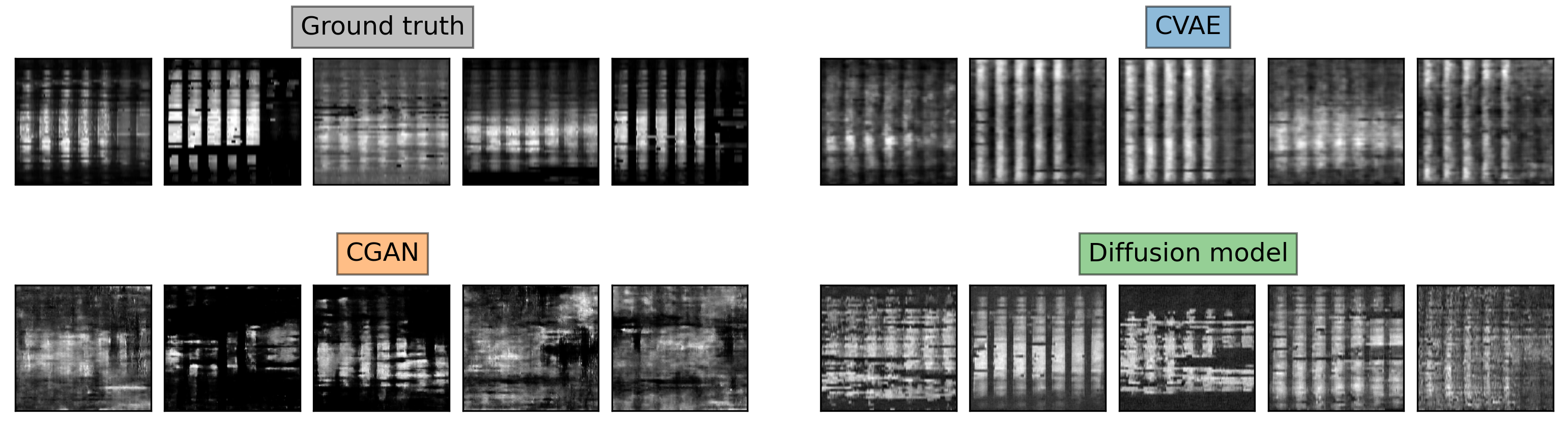}
        \label{fig:office_hotwater}
    \end{subfigure}
    
    \vspace{-1.5em}  
    
    \begin{subfigure}[b]{0.85\textwidth}
        \caption{Real data and generated data of \textbf{Steam meters} in \textbf{Education buildings}}
        \includegraphics[width=\textwidth]{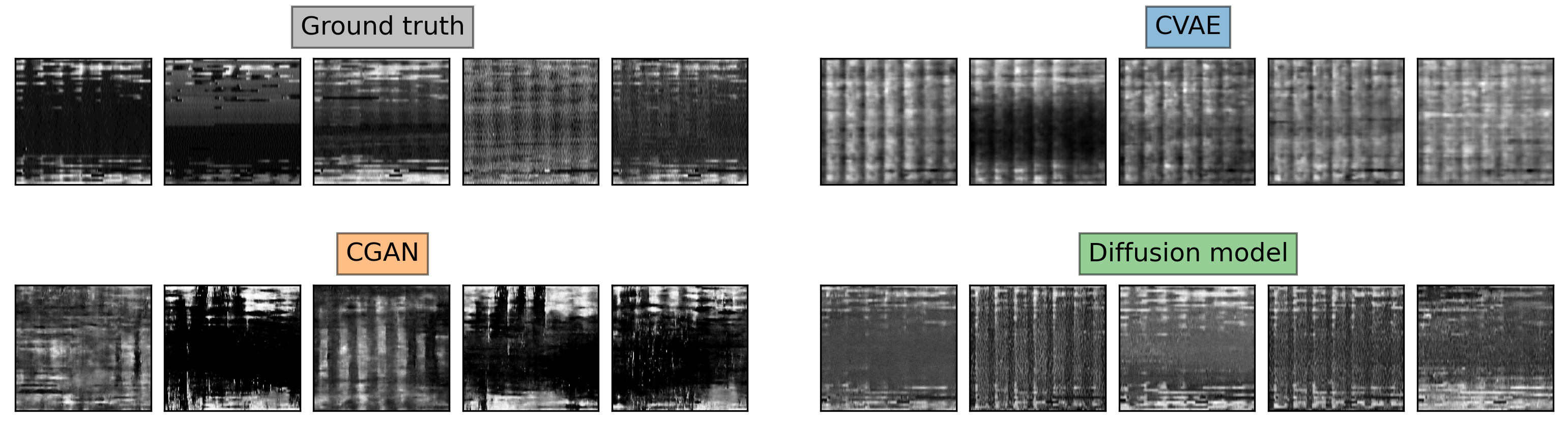}
        \label{fig:office_hotwater}
    \end{subfigure}
    
    \caption{Comparison of generated ground truth and two-dimensional heatmaps across models for different meter and building types.}
    \label{fig:clustered_heatmap_plots}
\end{figure*}

Upon closely examining Figures \ref{fig:clustered_ts_plots_annual},  \ref{fig:results_comp_plots} and \ref{fig:clustered_heatmap_plots}, certain nuances in the generated results become evident. These figures provide a side-by-side comparison of real and synthetic energy data generated using CGAN, CVAE, and the diffusion model.

The time series plots presented in Figure \ref{fig:clustered_ts_plots_annual} and Figure \ref{fig:results_comp_plots} provide a comprehensive view of energy consumption patterns across different time scales, specifically on an annual and monthly basis, to offer varied observational resolutions. Each plot contains 10 to 20 samples and their average line with the given conditions of metadata (e.g., in Figure \ref{fig:clustered_ts_plots_annual} (a), each subplot includes 14 samples of time series with the condition of electricity meter in an office building). From these plots, we can observe that CVAE outputs results that are more stable and consistent but lack variability — it maintains similar regular patterns regardless of the meter or building types. In contrast, CGAN offers better variability and can generate patterns closer to the ground truth based on the given condition, such as the seasonality between chilled water and steam meters, but its generated quality appears unstable with more fluctuations. The diffusion model, on the other hand, manages to balance both generation quality and variability. It distinctly reveals the seasonal differences of various meter types, and one can also notice the lower electricity consumption in educational buildings between semesters.

After that, the heatmaps in Figure \ref{fig:clustered_heatmap_plots} show us energy patterns in a two-dimensional heatmap view, enhancing our understanding of recurring weekly energy trends and inherent seasonality. Each subplot displays five samples corresponding to designated metadata conditions. A side-by-side comparison of these heatmaps reveals that CVAE provides the most consistent and smooth generated outputs. Conversely, CGAN exhibits pronounced more variability but with discernible distortions in the heatmap image. The proposed diffusion model stands out, delivering heatmaps that are both genuine in representation and rich in variability. Even the detailed changes in energy consumption on specific days and weeks are evident in the outputs generated by the diffusion model.

    \label{fig:clustered_ts_plots_monthly}

\section{Discussion}
Based on the methods and findings presented in this research, this section delves deeper into several aspects of synthetic energy data generation. These include its application potential, comparison of models' characteristics, future possibilities, and the limitations of the proposed approach.

\subsection{Meta-driven paradigm shift in energy generative models and its application potential}

The meta-driven approach introduced in this study fundamentally changes the paradigm of energy prediction. Our innovative methodology incorporates rich metadata, such as building and meter types, into the model architecture. This represents a departure from previous research in energy data generation, which predominantly focused on short-term generation and lacked the integration of metadata. The inclusion of metadata in our approach significantly enhances model personalization and specificity. Rather than providing generalized results, our model can generate tailored projections aligned with building-specific information, and it can generate data for extended periods, up to one year. The meta-driven philosophy shifts away from relying solely on historical data for targeted meters toward metadata-centric prediction.

The high-fidelity synthetic data produced by our approach unlocks diverse possibilities for building energy management. For instance, it enables comprehensive benchmarking by expanding available datasets. One can utilize the generated data for testing and validation before deploying changes. Moreover, considering the privacy concerns that may discourage utility companies from providing data, our proposed generative model has the potential to address this challenge. By designing the generative model to utilize non-sensitive metadata as inputs, such as building and meter types, we can fully harness the raw meter data for training purposes, thereby offering synthetic data as an alternative to directly exposing personal meter usage to the public.

\subsection{Comparative analysis of model performance and computational costs}

Our quantitative analysis clearly demonstrates the superiority of the proposed diffusion model over baseline approaches like CGAN and CVAE in terms of both generation diversity and fidelity. While CGAN provides good variability, it struggles with fidelity issues such as mode collapse. Conversely, CVAE yields stable predictions but tends to overly smooth them, resulting in insufficient diversity. These outcomes underscore the advantages of a metadata-conditioned diffusion framework for energy data synthesis.

However, in addition to the performance comparison, it is crucial to consider the computational cost of model training. In terms of computational requirements, the CVAE model has the lowest cost, taking approximately 10 minutes for training on a standard GPU machine in a Google Colaboratory (also known as Colab) environment. This efficiency stems from its simple encoder-decoder structure. 
CGAN, on the other hand, is more resource-intensive. Due to the adversarial training process, it requires around one hour to achieve convergence. 
In contrast, our diffusion model entails the highest cost, at approximately 6 hours for training. However, the stability and sample quality gains offered by the diffusion model may outweigh this cost. Overall, although our approach incurs a higher upfront computational cost, the long-term benefits in terms of performance and stability likely justify the one-time computational requirements.

\subsection{Future possibilities: Enhanced contextual considerations and prompt-driven generative models}

While this study focused on metadata like meter and building types, the model could be enhanced by incorporating additional contextual data sources. Details on a building's construction materials, insulation, occupancy patterns, and installed systems could provide further useful constraints. Geographic and climate data related to the building's location are also valuable in indicating energy demands tied to weather. Furthermore, electricity pricing schemes could offer insight into patterns affected by tariff structures. The modularity and extensibility of our proposed framework allow easy integration of supplementary context sources, progressively enhancing the model's awareness of the nuances that shape energy consumption profiles.

Beyond the realm of meta-driven generative models, an intriguing possibility is transitioning towards prompt-based frameworks that accept descriptive natural language prompts. Rather than feeding predefined attributes like meter types, users could provide elaborate prompts encapsulating rich contextual details and specifications for the energy data to be generated. This prompt-engineering approach would offer more flexibility and personalization. Exploring state-of-the-art prompt-based diffusion models could pave the path toward an era of customizable, user-centric energy data synthesis aligned with personalized prompts. This would represent the next evolutionary leap - from meta-driven to prompt-driven generative paradigms tailored for the energy domain.

\subsection{Limitations of the proposed method}
Despite the significant contributions of this study, it acknowledges several inherent limitations. First, regarding the model input, the absence of weather data as a condition for generation restricts its scope, especially when taking into account the strong correlation between weather and energy consumption. Furthermore, the potential to replace input metadata with natural language for enhanced flexibility and detailed descriptions is intriguing. For instance, a request like ``provide the electricity meter for a school building that is closed during non-term times and has concentrated occupancy from 8:00 to 18:00'' warrants further investigation. 
Secondly, the selected metadata in this research only represents a subset of all the available metadata. The ongoing pursuit involves integrating a more comprehensive set of metadata or seeking influential features that more accurately differentiate energy behaviors. 
Thirdly, the impact of generated energy data on subsequent tasks has not been tested. For instance, will the synthetic energy data enhance predictive performance?
Lastly, this research and previous studies have fixed generation lengths, focusing mainly on yearly or daily profiles. The capability to generate data over user-specified lengths would diversify application scenarios and is certainly worth exploring.

\section{Conclusion}

This research pioneered a novel meta-driven generative modeling approach to synthesize high-fidelity, long-term building energy consumption data. We implemented and evaluated three conditional generative models, namely Conditional VAE, Conditional GAN, and a tailored diffusion model. Quantitative analysis and visual inspection of synthetic meter data generated using the BDG2 dataset demonstrate the superiority of our proposed diffusion model in accurately capturing intricate statistical properties and temporal dynamics. Specifically, the model excels in both diversity and fidelity among competing generative models, achieving notable reductions of 36\% and 13\% in FID score and KL divergence, respectively, compared to the second-best performing model. Our contextual conditioning framework seamlessly integrates valuable metadata like meter and building types to produce customizable energy load profiles aligned with real-world constraints. This data synthesis capability could help overcome hurdles like data scarcity and privacy risks that impede effective energy management. The generated data can potentially empower a multitude of initiatives, including simulation, forecasting, diagnostics, and planning for buildings. This study establishes meta-driven generative modeling of energy data as a promising direction. Future work should explore additional sources of contextual data and scale the approach to district and urban scales. The proposed data synthesis strategy could catalyze transformative changes, enabling smarter energy utilization in buildings, communities, and cities.

\section{Reproducibility}
This analysis can be reproduced using the data and code from the following GitHub repository: \url{https://github.com/buds-lab/energy-diffusion}. 

\section*{Funding}
This research is funded by the NUS-based Singapore MOE Tier 1 Grant titled Ecological Momentary Assessment (EMA) for Built Environment Research (A-0008301-01-00).

\section*{CRediT author statement}
\textbf{CF}: Conceptualization, Methodology, Software, Formal Analysis, Investigation, Data Curation, Visualization, Writing - Original Draft; 
\textbf{HK}: Methodology, Supervision, Writing - Reviewing \& Editing; 
\textbf{MQ}: Methodology, Writing - Reviewing \& Editing; 
\textbf{CM}: Conceptualization, Methodology, Resources, Writing - Reviewing \& Editing, Supervision, Project administration, Funding acquisition.

\bibliography{mybibfile}

\begin{thebibliography}{10}
\expandafter\ifx\csname url\endcsname\relax
  \def\url#1{\texttt{#1}}\fi
\expandafter\ifx\csname urlprefix\endcsname\relax\def\urlprefix{URL }\fi
\expandafter\ifx\csname href\endcsname\relax
  \def\href#1#2{#2} \def\path#1{#1}\fi

\bibitem{creutzig2015global}
F.~Creutzig, G.~Baiocchi, R.~Bierkandt, P.-P. Pichler, K.~C. Seto, Global typology of urban energy use and potentials for an urbanization mitigation wedge, Proceedings of the national academy of sciences 112~(20) (2015) 6283--6288.

\bibitem{stephenson2010energy}
J.~Stephenson, B.~Barton, G.~Carrington, D.~Gnoth, R.~Lawson, P.~Thorsnes, Energy cultures: A framework for understanding energy behaviours, Energy policy 38~(10) (2010) 6120--6129.

\bibitem{sovacool2021global}
B.~K. Sovacool, A.~Hook, S.~Sareen, F.~W. Geels, Global sustainability, innovation and governance dynamics of national smart electricity meter transitions, Global Environmental Change 68 (2021) 102272.

\bibitem{arora2022review}
S.~Arora, A.~Thakur, A.~Singh, S.~Rana, D.~Singh, A review on smart energy meters and their market trends, in: 2022 International Conference on Emerging Trends in Engineering and Medical Sciences (ICETEMS), IEEE, 2022, pp. 167--172.

\bibitem{union2009directive}
E.~Union, Directive 2009/28/ec of the european parliament and of the council of 23 april 2009 on the promotion of the use of energy from renewable sources and amending and subsequently repealing directives 2001/77/ec and 2003/30/ec, Official Journal of the European Union 5 (2009) 2009.

\bibitem{cooper2021electric}
A.~Cooper, M.~Shuster, J.~Lash, Electric company smart meter deployments: foundation for a smart grid (2021 update), Institue for Electric Innovation: Washington, DC, USA.

\bibitem{mcdaniel2009security}
P.~McDaniel, S.~McLaughlin, Security and privacy challenges in the smart grid, IEEE security \& privacy 7~(3) (2009) 75--77.

\bibitem{balta2013social}
N.~Balta-Ozkan, R.~Davidson, M.~Bicket, L.~Whitmarsh, Social barriers to the adoption of smart homes, Energy policy 63 (2013) 363--374.

\bibitem{langer2013privacy}
L.~Langer, F.~Skopik, G.~Kienesberger, Q.~Li, Privacy issues of smart e-mobility, in: IECON 2013-39th Annual Conference of the IEEE Industrial Electronics Society, IEEE, 2013, pp. 6682--6687.

\bibitem{kazmi2021towards}
H.~Kazmi, {\'I}.~Munn{\'e}-Collado, F.~Mehmood, T.~A. Syed, J.~Driesen, Towards data-driven energy communities: A review of open-source datasets, models and tools, Renewable and Sustainable Energy Reviews 148 (2021) 111290.

\bibitem{kazmi2023ten}
H.~Kazmi, C.~Fu, C.~Miller, Ten questions concerning data-driven modelling and forecasting of operational energy demand at building and urban scale, Building and Environment 239 (2023) 110407.

\bibitem{somu2021deep}
N.~Somu, G.~R. MR, K.~Ramamritham, A deep learning framework for building energy consumption forecast, Renewable and Sustainable Energy Reviews 137 (2021) 110591.

\bibitem{pham2020predicting}
A.-D. Pham, N.-T. Ngo, T.~T.~H. Truong, N.-T. Huynh, N.-S. Truong, Predicting energy consumption in multiple buildings using machine learning for improving energy efficiency and sustainability, Journal of Cleaner Production 260 (2020) 121082.

\bibitem{olu2022building}
R.~Olu-Ajayi, H.~Alaka, I.~Sulaimon, F.~Sunmola, S.~Ajayi, Building energy consumption prediction for residential buildings using deep learning and other machine learning techniques, Journal of Building Engineering 45 (2022) 103406.

\bibitem{dolara2017weather}
A.~Dolara, A.~Gandelli, F.~Grimaccia, S.~Leva, M.~Mussetta, Weather-based machine learning technique for day-ahead wind power forecasting, in: 2017 IEEE 6th international conference on renewable energy research and applications (ICRERA), IEEE, 2017, pp. 206--209.

\bibitem{gensler2016deep}
A.~Gensler, J.~Henze, B.~Sick, N.~Raabe, Deep learning for solar power forecasting—an approach using autoencoder and lstm neural networks, in: 2016 IEEE international conference on systems, man, and cybernetics (SMC), IEEE, 2016, pp. 002858--002865.

\bibitem{tian2022developing}
C.~Tian, T.~Niu, W.~Wei, Developing a wind power forecasting system based on deep learning with attention mechanism, Energy 257 (2022) 124750.

\bibitem{hafeez2020novel}
G.~Hafeez, K.~S. Alimgeer, Z.~Wadud, Z.~Shafiq, M.~U. Ali~Khan, I.~Khan, F.~A. Khan, A.~Derhab, A novel accurate and fast converging deep learning-based model for electrical energy consumption forecasting in a smart grid, Energies 13~(9) (2020) 2244.

\bibitem{avalos2020comparative}
E.~E. Avalos, M.~R. Licea, H.~R. Gonz{\'a}lez, A.~E. Calder{\'o}n, A.~B. Guti{\'e}rrez, F.~P. Pinal, Comparative analysis of multivariable deep learning models for forecasting in smart grids, in: 2020 IEEE International Autumn Meeting on Power, Electronics and Computing (ROPEC), Vol.~4, IEEE, 2020, pp. 1--6.

\bibitem{hafeez2020electric}
G.~Hafeez, K.~S. Alimgeer, I.~Khan, Electric load forecasting based on deep learning and optimized by heuristic algorithm in smart grid, Applied Energy 269 (2020) 114915.

\bibitem{fumo2015regression}
N.~Fumo, M.~R. Biswas, Regression analysis for prediction of residential energy consumption, Renewable and sustainable energy reviews 47 (2015) 332--343.

\bibitem{ahmad2020review}
T.~Ahmad, H.~Chen, A review on machine learning forecasting growth trends and their real-time applications in different energy systems, Sustainable Cities and Society 54 (2020) 102010.

\bibitem{Miller2022-jj}
C.~Miller, L.~Hao, C.~Fu, Gradient boosting machines and careful pre-processing work best: {ASHRAE} great energy predictor {III} lessons learned, in: ASHRAE Transactions, Vol. 128, ASHRAE, 2022, pp. 405--413.

\bibitem{kazmi2019multi}
H.~Kazmi, J.~Suykens, A.~Balint, J.~Driesen, Multi-agent reinforcement learning for modeling and control of thermostatically controlled loads, Applied energy 238 (2019) 1022--1035.

\bibitem{amasyali2018review}
K.~Amasyali, N.~M. El-Gohary, A review of data-driven building energy consumption prediction studies, Renewable and Sustainable Energy Reviews 81 (2018) 1192--1205.

\bibitem{hensen2012building}
J.~L. Hensen, R.~Lamberts, Building performance simulation for design and operation, Routledge, 2012.

\bibitem{hong2020generation}
T.~Hong, D.~Macumber, H.~Li, K.~Fleming, Z.~Wang, Generation and representation of synthetic smart meter data, in: Building Simulation, Vol.~13, Springer, 2020, pp. 1205--1220.

\bibitem{reddy2006literature}
T.~A. Reddy, Literature review on calibration of building energy simulation programs: Uses, problems, procedures, uncertainty, and tools., ASHRAE transactions 112~(1).

\bibitem{fabrizio2015}
E.~Fabrizio, V.~Monetti, Methodologies and {{Advancements}} in the {{Calibration}} of {{Building Energy Models}}, Energies 8~(4) (2015) 2548--2574.
\newblock \href {http://dx.doi.org/10.3390/en8042548} {\path{doi:10.3390/en8042548}}.

\bibitem{goodfellow2014generative}
I.~Goodfellow, J.~Pouget-Abadie, M.~Mirza, B.~Xu, D.~Warde-Farley, S.~Ozair, A.~Courville, Y.~Bengio, Generative adversarial nets, Advances in neural information processing systems 27.

\bibitem{kingma2013auto}
D.~P. Kingma, M.~Welling, Auto-encoding variational bayes, arXiv preprint arXiv:1312.6114.

\bibitem{ho2020denoising}
J.~Ho, A.~Jain, P.~Abbeel, Denoising diffusion probabilistic models, Advances in neural information processing systems 33 (2020) 6840--6851.

\bibitem{yeEnergyBuildingsEvaluating2022}
Y.~Ye, M.~Strong, Y.~Lou, C.~A. Faulkner, W.~Zuo, S.~Upadhyaya, Energy \& {{Buildings Evaluating}} performance of different generative adversarial networks for large-scale building power demand prediction, Energy \& Buildings 269 (2022) 112247.
\newblock \href {http://dx.doi.org/10.1016/j.enbuild.2022.112247} {\path{doi:10.1016/j.enbuild.2022.112247}}.

\bibitem{chen2018model}
Y.~Chen, Y.~Wang, D.~Kirschen, B.~Zhang, Model-free renewable scenario generation using generative adversarial networks, IEEE Transactions on Power Systems 33~(3) (2018) 3265--3275.

\bibitem{el2020data}
S.~El~Kababji, P.~Srikantha, A data-driven approach for generating synthetic load patterns and usage habits, IEEE Transactions on Smart Grid 11~(6) (2020) 4984--4995.

\bibitem{wang2020generating}
Z.~Wang, T.~Hong, Generating realistic building electrical load profiles through the generative adversarial network (gan), Energy and Buildings 224 (2020) 110299.

\bibitem{yan2020generative}
K.~Yan, A.~Chong, Y.~Mo, Generative adversarial network for fault detection diagnosis of chillers, Building and Environment 172 (2020) 106698.

\bibitem{fu2023enhancing}
C.~Fu, H.~Kazmi, M.~Quintana, C.~Miller, Enhancing classification of energy meters with limited labels using a semi-supervised generative model, in: Proceedings of the 10th ACM International Conference on Systems for Energy-Efficient Buildings, Cities, and Transportation, 2023, pp. 450--453.

\bibitem{quintana2020balancing}
M.~Quintana, S.~Schiavon, K.~W. Tham, C.~Miller, Balancing thermal comfort datasets: We gan, but should we?, in: Proceedings of the 7th ACM International Conference on Systems for Energy-Efficient Buildings, Cities, and Transportation, 2020, pp. 120--129.

\bibitem{mirza2014conditional}
M.~Mirza, S.~Osindero, Conditional generative adversarial nets, arXiv preprint arXiv:1411.1784.

\bibitem{ramponi2018t}
G.~Ramponi, P.~Protopapas, M.~Brambilla, R.~Janssen, T-cgan: Conditional generative adversarial network for data augmentation in noisy time series with irregular sampling, arXiv preprint arXiv:1811.08295.

\bibitem{fu2019time}
R.~Fu, J.~Chen, S.~Zeng, Y.~Zhuang, A.~Sudjianto, Time series simulation by conditional generative adversarial net, arXiv preprint arXiv:1904.11419.

\bibitem{guo2018long}
J.~Guo, S.~Lu, H.~Cai, W.~Zhang, Y.~Yu, J.~Wang, Long text generation via adversarial training with leaked information, in: Proceedings of the AAAI conference on artificial intelligence, Vol.~32, 2018.

\bibitem{baasch2021conditional}
G.~Baasch, G.~Rousseau, R.~Evins, A conditional generative adversarial network for energy use in multiple buildings using scarce data, Energy and AI 5 (2021) 100087.

\bibitem{fochesato2022use}
M.~Fochesato, F.~Khayatian, D.~F. Lima, Z.~Nagy, On the use of conditional timegan to enhance the robustness of a reinforcement learning agent in the building domain, in: Proceedings of the 9th ACM International Conference on Systems for Energy-Efficient Buildings, Cities, and Transportation, 2022, pp. 208--217.

\bibitem{salatiello2023synthesizing}
A.~Salatiello, Y.~Wang, G.~Wichern, T.~Koike-Akino, Y.~Ohta, Y.~Kaneko, C.~Laughman, A.~Chakrabarty, Synthesizing building operation data with generative models: Vaes, gans, or something in between?, in: Companion Proceedings of the 14th ACM International Conference on Future Energy Systems, 2023, pp. 125--133.

\bibitem{nuastuasescu2022conditional}
G.-S. N{\u{a}}st{\u{a}}sescu, D.-C. Cercel, Conditional wasserstein gan for energy load forecasting in large buildings, in: 2022 International Joint Conference on Neural Networks (IJCNN), IEEE, 2022, pp. 1--8.

\bibitem{park2018data}
N.~Park, M.~Mohammadi, K.~Gorde, S.~Jajodia, H.~Park, Y.~Kim, Data synthesis based on generative adversarial networks, arXiv preprint arXiv:1806.03384.

\bibitem{croitoru2023diffusion}
F.-A. Croitoru, V.~Hondru, R.~T. Ionescu, M.~Shah, Diffusion models in vision: A survey, IEEE Transactions on Pattern Analysis and Machine Intelligence.

\bibitem{fu2022using}
C.~Fu, C.~Miller, Using google trends as a proxy for occupant behavior to predict building energy consumption, Applied Energy 310 (2022) 118343.

\bibitem{miller2020building}
C.~Miller, A.~Kathirgamanathan, B.~Picchetti, P.~Arjunan, J.~Y. Park, Z.~Nagy, P.~Raftery, B.~W. Hobson, Z.~Shi, F.~Meggers, The building data genome project 2, energy meter data from the ashrae great energy predictor iii competition, Scientific data 7~(1) (2020) 368.

\bibitem{miller2020ashrae}
C.~Miller, P.~Arjunan, A.~Kathirgamanathan, C.~Fu, J.~Roth, J.~Y. Park, C.~Balbach, K.~Gowri, Z.~Nagy, A.~D. Fontanini, et~al., The ashrae great energy predictor iii competition: Overview and results, Science and Technology for the Built Environment 26~(10) (2020) 1427--1447.

\bibitem{sohn2015learning}
K.~Sohn, H.~Lee, X.~Yan, Learning structured output representation using deep conditional generative models, Advances in neural information processing systems 28.

\bibitem{bao2017cvae}
J.~Bao, D.~Chen, F.~Wen, H.~Li, G.~Hua, Cvae-gan: fine-grained image generation through asymmetric training, in: Proceedings of the IEEE international conference on computer vision, 2017, pp. 2745--2754.

\bibitem{odena2017conditional}
A.~Odena, C.~Olah, J.~Shlens, Conditional image synthesis with auxiliary classifier gans, in: International conference on machine learning, PMLR, 2017, pp. 2642--2651.

\bibitem{che2016mode}
T.~Che, Y.~Li, A.~P. Jacob, Y.~Bengio, W.~Li, Mode regularized generative adversarial networks, arXiv preprint arXiv:1612.02136.

\bibitem{kushwaha2020study}
V.~Kushwaha, G.~Nandi, et~al., Study of prevention of mode collapse in generative adversarial network (gan), in: 2020 IEEE 4th Conference on Information \& Communication Technology (CICT), IEEE, 2020, pp. 1--6.

\bibitem{dhariwal2021diffusion}
P.~Dhariwal, A.~Nichol, Diffusion models beat gans on image synthesis, Advances in neural information processing systems 34 (2021) 8780--8794.

\bibitem{borji2019pros}
A.~Borji, Pros and cons of gan evaluation measures, Computer vision and image understanding 179 (2019) 41--65.

\bibitem{theis2015note}
L.~Theis, A.~v.~d. Oord, M.~Bethge, A note on the evaluation of generative models, arXiv preprint arXiv:1511.01844.

\bibitem{heusel2017gans}
M.~Heusel, H.~Ramsauer, T.~Unterthiner, B.~Nessler, S.~Hochreiter, Gans trained by a two time-scale update rule converge to a local nash equilibrium, Advances in neural information processing systems 30.

\bibitem{kullback1951information}
S.~Kullback, R.~A. Leibler, On information and sufficiency, The annals of mathematical statistics 22~(1) (1951) 79--86.

\end{thebibliography}

\end{document}